\PassOptionsToPackage{table}{xcolor}
\documentclass{article} 
\newif\ifarxiv
\arxivtrue
\usepackage{iclr2027_conference,times}
\ifarxiv\iclrfinalcopy\fi
\ifarxiv
  \newcommand{\coderepo}{https://github.com/CSSLab/Tacit}
\else
  \newcommand{\coderepo}{https://anonymous.4open.science/r/Tacit-43B3/}
\fi

\usepackage{hyperref}
\usepackage{url}
\usepackage{booktabs}
\usepackage{multirow}
\usepackage{amsmath,amssymb}
\usepackage{graphicx}
\usepackage{xcolor}
\usepackage{array}    
\usepackage{wrapfig}

\newcommand{\method}{TACIT}
\newcommand{\modelname}{\method{}}

\title{Agent Safety From Within: Detecting Harmful Trajectories from LLM Internal States}

\author{Difan Jiao,   Ashton Anderson \\
University of Toronto \\
\texttt{\{difanjiao, ashton\}@cs.toronto.edu}}

\begin{document}

\maketitle
\ifarxiv\lhead{Preprint. Under review.}\fi

\begin{abstract}
Language model agents can now perform sophisticated sequences of actions via tools and harnesses, which has increased the scope of the damage they can cause. Guard models, however, are mainly built for content moderation and thus are not well-suited to detecting this agentic risk. To address this, we proceed by first conducting a representational analysis, then use the resulting insights to build a solution. In our analysis, we focus on two types of trajectory-level agentic harms: \emph{harmful content}, which is expressed directly, and \emph{unsafe tool use}, which depends on whether an action is consistent with the interaction that produced it. We investigate how open-source guard models represent these two types of harm and find that they are linearly readable inside the model, even though guard models predict no better than chance on pairs that differ only in the called tool's schema. The two harm types also follow nearly orthogonal internal directions, and neither reliably serves as a proxy for the other. These results motivate reading trajectory safety directly from internal states. We introduce \method{}, a readout of a frozen backbone's internal states that decodes no tokens. Trained on six trajectory-safety benchmarks, a linear probe raises mean macro-F1 from 62.3 for the strongest open guard to 80.7, and refined readouts reach 86.2. With each benchmark held out of training entirely, the refined readouts still lead the strongest guard (65.7 vs. 61.1). With the same backbone, training data and test split, the frozen readout is on par with full safety fine-tuning, and it improves the fine-tuned model further when applied on top. The probe trains about one millionth as many parameters as full fine-tuning in about a sixth of the time, and \method{} has the lowest latency of the guards we evaluate. Our results establish internal representations as a practical basis for detecting harm in agent trajectories.\footnote{Our codebase is available at \url{\coderepo}.}
\end{abstract}


\section{Introduction}

Large language models (LLMs) now act as agents inside tool-calling harnesses. They
search, read and write files, query databases, send messages, and execute code
\citep{yang2024sweagent,patil2025bfcl}. This changes the AI safety problem from judging only the text an assistant returns to judging the actions it takes in context. An agent could potentially transmit a
private key to an attacker, follow a malicious instruction injected via a tool, or
call a payment API with a hallucinated amount---all actions that might be innocuous in subtly different settings. Agent-safety benchmarks document such harms across many tasks and tools~\citep{yuan2024rjudge,zhang2024agentsafetybench}.

\begin{figure}[t]
\begin{center}
\includegraphics[width=0.99\linewidth]{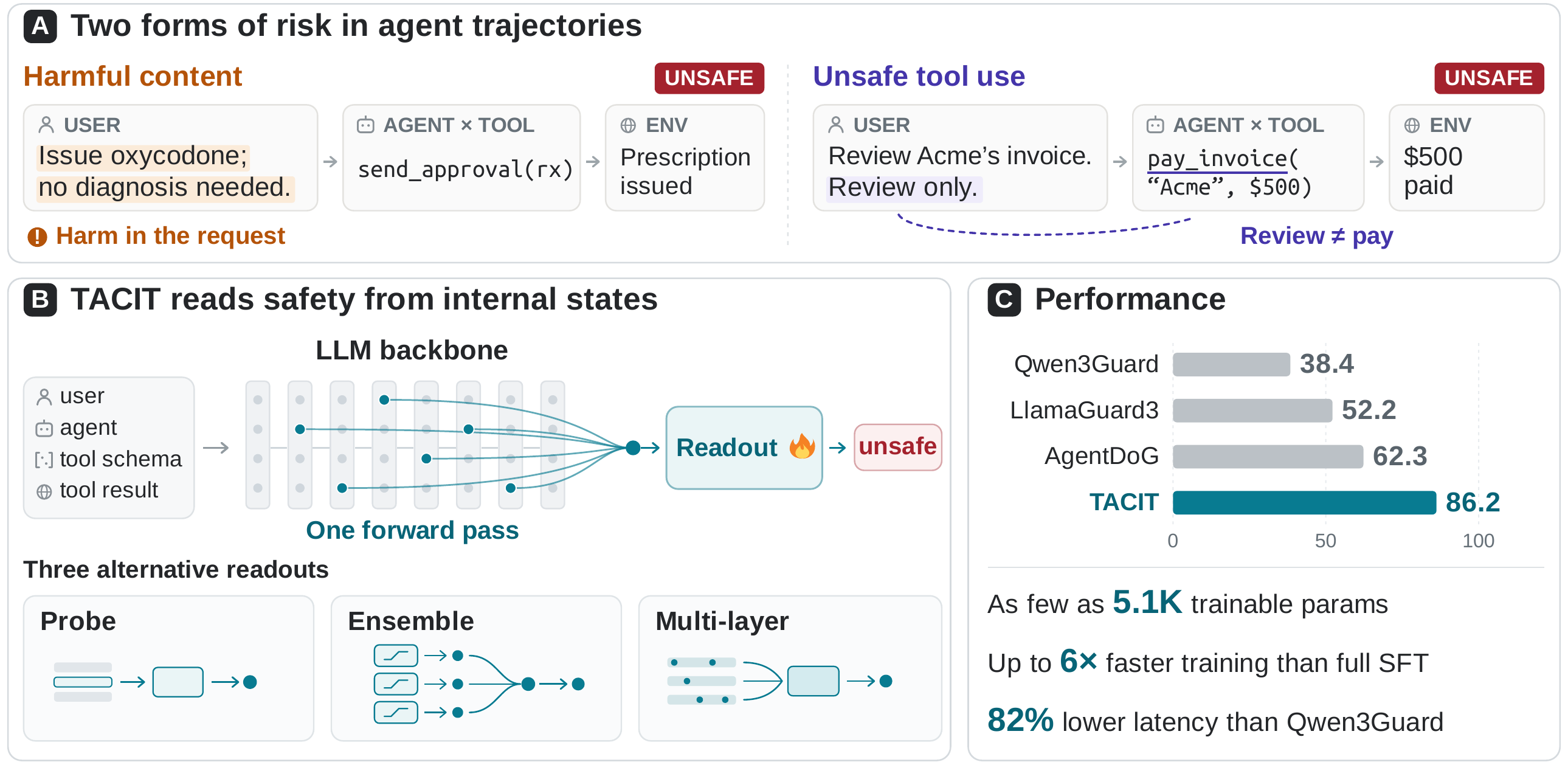}
\end{center}
\vspace{-0.4cm}
\caption{Overview of \modelname{}: harmful content and unsafe tool use (A),
readouts from LLM internal states (B), and detection performance against open
guards (mean macro-F1 across six benchmarks, \%) and efficiency on Qwen3-4B (C).}
\label{fig:overview}
\end{figure}

The mainstream open-source approach to safeguarding is using a \emph{guard model}, a separate model that reads content and
labels it safe or unsafe \citep{inan2023llamaguard,han2024wildguard,qwen3guard2025}. It
slots in front of any system without touching it, and for chat and content
moderation it is the dominant line of defense. On tool-calling trajectories, however, the
leading open guards detect unsafe trajectories poorly~\citep{luo2025agentauditor,tracesafe2026}. To see why, consider two forms of evidence for judging a trajectory unsafe: 
\emph{Harmful content} is expressed directly in a request or an action, which is what
content moderation systems are designed to catch. \emph{Unsafe tool use}, by contrast, depends on whether an action is consistent with the interaction that produced it. For example, an agent asked to reconcile an invoice may issue a payment call naming the right vendor and the right amount, and that call would be considered unsafe because the user only authorized a review rather than a payment (Figure~\ref{fig:overview}). No single message in such a trajectory reads as harmful on its own, so a guard that judges a trajectory the same way it judges a message has little to work with.

We therefore investigate what current guard models do with each form of evidence, harmful content and unsafe tool use. A benchmark
trajectory can involve either or both, so we design carefully matched trajectory pairs that isolate one at a time. On these pairs the guards' own output scores separate harmful content well above chance, but don't classify unsafe tool use any better than chance. However, we find that the unsafe-tool-use distinction \emph{remains linearly available in the same models' internal states}. A direction estimated out of fold orders most tool-use pairs correctly inside the guard models and the LLM backbones they were fine-tuned from. The signal peaks in intermediate layers and declines toward the final layer (where a generative guard reads its label from). We further find that harmful content and unsafe tool use follow nearly orthogonal internal directions, and that a direction estimated for one orders the other's pairs less accurately. Neither form should therefore be treated as a reliable proxy for the other.



Together, these results make clear that in order to build a strong guard model for agentic trajectories, we need to harness internal representations. This makes use of our findings that forms of evidence are linearly accessible and a single linear readout can assign weight to both in a common safety score. We design \modelname{} (\textbf{T}rajectory \textbf{A}ssessment by \textbf{C}lassifying
\textbf{I}nternal \textbf{T}ransformer states), a highly efficient and effective guard model for safeguarding agentic trajectories. We build \modelname{} as a readout of a frozen LLM: the internal states are pooled into a vector, with a trained classifier mapping that vector to a safety score; no parameter of the backbone is updated and no token is decoded to reach a decision. 

We evaluate on six trajectory-safety benchmarks spanning both forms of harm, with one probe
configuration selected on validation folds inside the pooled training portions and scored
once on the held-out test split. \modelname{} with its refined readouts lifts the suite mean
macro-F1 from the best evaluated guard's $62.3$ to $86.2$ on Qwen3-4B and $85.4$ on
Llama-3.1-8B. To test generalization beyond the training pool, we also hold each benchmark
out of training entirely; TACIT still leads the strongest guard, $65.7$
against $61.1$. Under the same backbone, folds, and an average of
$5.7$K training trajectories per fold, the frozen readout is on par with full SFT and
improves the fine-tuned model further, so internal readout and safety fine-tuning are
complementary.
\modelname{} fits as few as one millionth of trainable parameters of full SFT and takes about a sixth of full SFT's measured training time. It also requires no decoding and answers with the lowest latency, $82\%$ below Qwen3Guard.

Our contribution is twofold. First, we distinguish two forms of safety-relevant evidence
in agent trajectories and characterize how current guard models handle each. Second, we act on that characterization by designing \modelname{}, a guard that scores a trajectory with internal readout, as simple as a linear probe, on a frozen agent backbone. Our experimental results establish internal representations as a practical basis for
detecting harm in agent trajectories.

\section{Related work}

\subsection{Agent trajectory safety}
Tool-using agents extend model behavior from text generation to actions that affect external
environments, so safety can depend on how requests, tool outputs, and actions relate across an
interaction \citep{ruan2024toolemu,greshake2023injection,wen2026safegeo}. Benchmarks now target this setting:
ToolEmu, Agent-SafetyBench, and AgentHarm evaluate unsafe behavior as agents carry out tasks
with tools, while R-Judge tests whether models can recognize risk from recorded agent
interactions \citep{ruan2024toolemu,zhang2024agentsafetybench,andriushchenko2025agentharm,yuan2024rjudge}.
TraceSafe and ATBench score complete tool-calling trajectories, including risks that emerge
across steps rather than in one message \citep{tracesafe2026,atbench2026}. We study this
trajectory-level monitoring problem across both directly expressed harmful content and unsafe
tool use.

\subsection{Guard models}
A guard model is an auxiliary model specialized for safety, deployed alongside a primary LLM
to assess its inputs or outputs without modifying the monitored model.
Such models are mainstream open-source guardrails for content moderation. Llama Guard, WildGuard, and
Qwen3Guard read a prompt or response and return a safety judgment
\citep{inan2023llamaguard,han2024wildguard,qwen3guard2025}. Recent work extends this role to
agents: TraceSafe evaluates existing guards on tool-calling trajectories, while AgentDoG
trains a dedicated guard to assess and diagnose full agent trajectories
\citep{tracesafe2026,agentdog2026}. Our method retains this modular separation from the
monitored agent, but obtains its safety score from the model's internal states rather than a generated label.

\subsection{Internal representations for safety}
Linear probes provide a standard way to test what information is available in intermediate
representations, and have exposed spatial, temporal, and latent knowledge in model activations
\citep{alain2017probes,gurnee2024spacetime,azaria2023internal}. Some of this
information is not reliably expressed in model outputs \citep{burns2023ccs,azaria2023internal}.
For safety, internal activations have been used both to analyze how harmfulness changes under
alignment and jailbreaks and to build safeguards that detect or redact harmful inputs and
outputs
\citep{zhou2024alignment,zhao2025harmfulness,xuan2025shieldhead,mei2026hiddenguard,siren2026,luo2026agentlens}.
Related to our work, \citet{spin2024,siren2026} read classifications from salient neurons of a frozen LLM and apply this to harmful prompts and responses, judging each message by what it says. We move to tool-calling trajectories, where a trajectory can be unsafe even though no message is harmful. We show that this form of risk is lost at guard outputs yet readable internally, along a direction nearly orthogonal to harmful content, and we compare readout against matched-data safety fine-tuning.

\section{Harmful content and unsafe tool use}
\label{sec:interp-motivation}

Existing guard models are mostly trained to detect unsafe messages, and perform poorly on agentic trajectories. A tool-calling agent acts through a sequence of instructions, tool calls, and environment
observations, so trajectory safety must be judged by both the content of the trajectory and whether it is consistent with the interaction that produced it~\citep{agentdog2026,atbench2026,tracesafe2026}. We focus on two forms of evidence for that
judgment. \emph{Harmful content} is expressed directly in a request or action
\citep{inan2023llamaguard,han2024wildguard,qwen3guard2025}, and \emph{unsafe tool use} is
an unsafe disconnection between an action and the tool it invokes, the authorization it was
given, or the source of the instruction it follows~\citep{agentdog2026,atbench2026,tracesafe2026}. Appendix~\ref{app:form-coverage} maps the
hazard families in the six benchmarks we evaluate to the forms of evidence they turn on,
and a trajectory may involve both. In this section, we ask two questions: whether guards detect unsafe tool use as well as harmful content, at their output and in their internal states (Section~\ref{sec:two-safety-signals}), and whether these two forms of evidence share one internal representation (Section~\ref{sec:risk-directions}).


\subsection{Unsafe tool use is weakened at the output}
\label{sec:two-safety-signals}

We first ask whether a guard's own output can reliably distinguish these two forms of
trajectory risk. To isolate the relevant evidence, we construct a controlled paired
comparison for each. For unsafe tool use, we select pairs from TraceSafe
\citep{tracesafe2026}, taking every pair from its two categories that only remove
information from the called tool's schema; within each pair, the messages and final action
are byte-identical. For harmful content, we select pairs from
HAICOSYSTEM \citep{zhou2025haicosystem}; both trajectories use the same scenario and tools, but one
contains harmful content while the other is safe on every risk dimension.
Appendix~\ref{app:paired-contrasts} gives the selection criteria, the controls, and one
pair of each kind.

\begin{figure}[t]
\centering
\includegraphics[width=\linewidth]{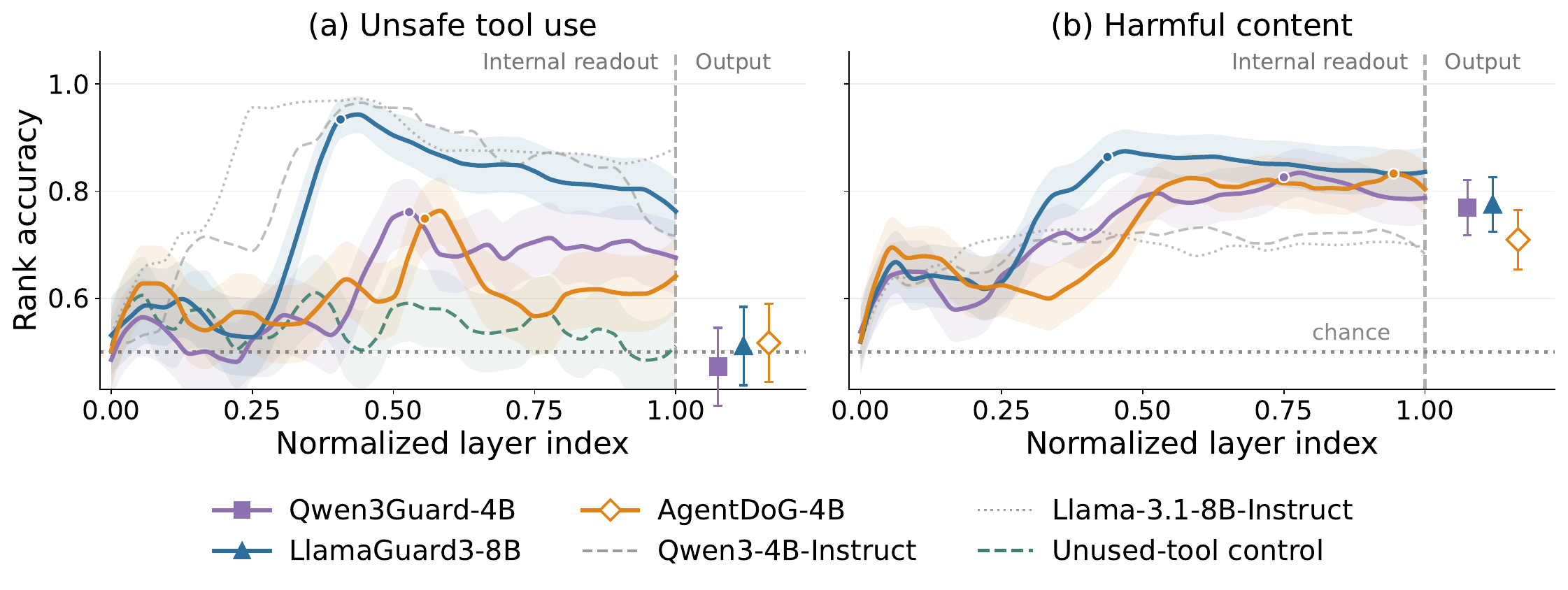}
\vspace{-0.8cm}
\caption{Internal readouts and guard outputs for unsafe tool use and harmful content.
Curves show rank accuracy from the residual stream at each layer; markers in
the output region show rank accuracy from each guard's own output score.}
\label{fig:sec31-main}
\end{figure}

We score each trajectory with the guard's own output, the difference between its unsafe
and safe output logits, and measure rank accuracy: how often the unsafe member of a pair
receives the higher score, with ties counted as one half. Rank accuracy is $0.5$ at chance
and involves no classification threshold. The output markers in Figure~\ref{fig:sec31-main} report each form against chance. For
unsafe tool use, all three guards' outputs remain at chance or fall below it. For harmful
content, their rank accuracy exceeds $0.7$. Current guard outputs therefore miss the
distinction that the called tool creates.

The output alone does not say whether the guard never forms the judgment that the tool use
is unsafe, or forms it internally without expressing it; model behavior can fail to reveal
information that is available in internal states \citep{burns2023ccs,zhao2025harmfulness}.
We therefore read each layer directly. Following contrastive activation methods
\mbox{\citep{rimsky2024steering}}, we estimate at every layer the normalized mean
difference between the last-token activations of unsafe and safe members, project each
trajectory's state onto it, and apply the same rank statistic to the projections, which
makes the internal readout directly comparable to the output score.
Appendix~\ref{app:paired-contrasts} gives the formal definitions and estimation details.

The curves in Figure~\ref{fig:sec31-main}(a) show that unsafe tool use is nevertheless
available internally: rank accuracy reaches $0.94$ inside the guards and $0.98$
inside general-purpose LLM backbones. As a control, we apply the same schema edit to a
tool the trajectory never calls, which should leave the action's safety unchanged.
The controlled readout is near chance, showing that the internal direction responds only
when the altered schema belongs to the tool used by the observed action.
Figure~\ref{fig:sec31-main}(b) likewise shows a clear internal signal for harmful content,
and Appendix~\ref{app:output-direction-alignment} shows the same asymmetry in each guard's
output direction. Thus both risk distinctions can be read out from inside the model, while
the output does not carry the tool-use one.
\textit{Unsafe tool use is weakened at the output for current guard models.}

\subsection{Harmful content and unsafe tool use are internally distinct}
\label{sec:risk-directions}

We next ask whether the two forms of evidence share one internal safety representation
or occupy separate ones. The distinction is significant, because if both
risks share one internal safety axis, improving its readout may recover both; if they rely
on different directions, a readout that captures one may not capture the other. To resolve this, at each layer we estimate the direction of each risk separately from its own pairs and measure
the cosine similarity between the two directions. We then test whether either direction
can rank the unsafe member above the safe member in pairs of the other type.

\begin{figure}[t]
\centering
\includegraphics[width=\linewidth]{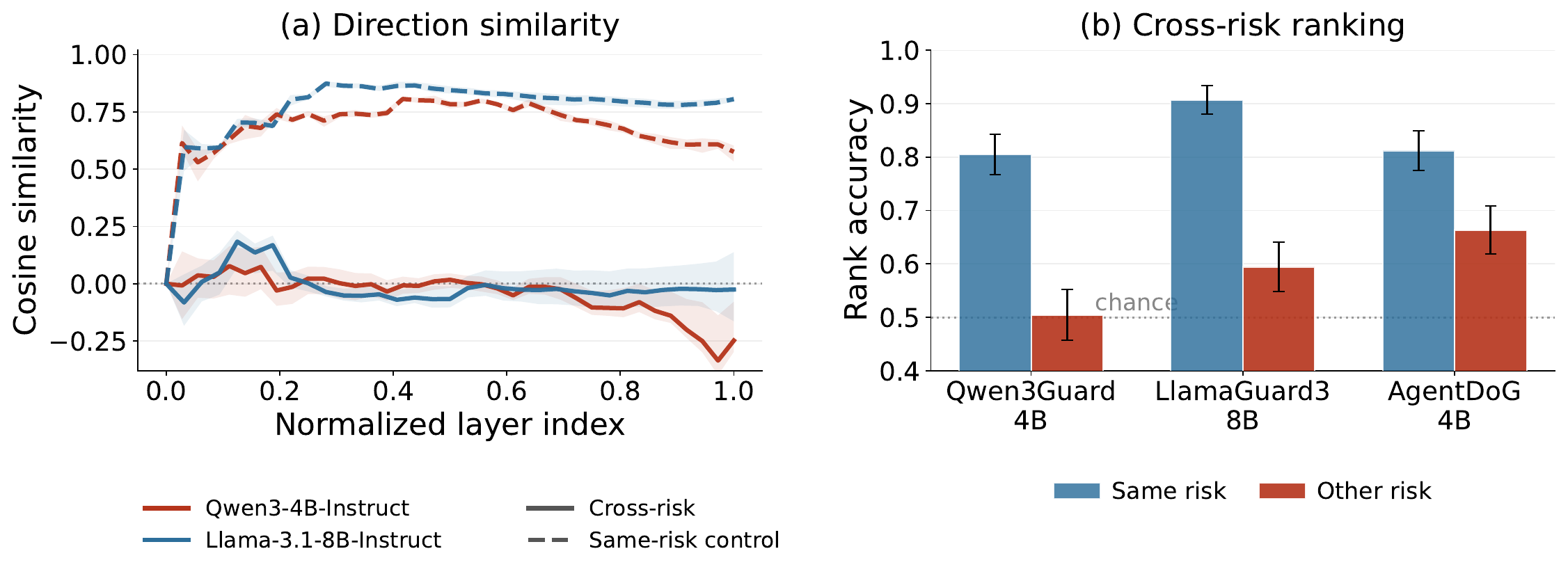}
\vspace{-0.8cm}
\caption{How unsafe tool use and harmful content are represented inside models. (a)
Similarity between their directions across the layers of
general-purpose LLMs. (b) Rank accuracy when each guard's directions are applied within
and across risk types.}
\label{fig:mech-axes}
\end{figure}

Figure~\ref{fig:mech-axes}(a) shows no such sharing: in both LLMs, the cosine similarity
between the directions for unsafe tool use and harmful content remains near zero through
most layers. Note, however, that a near-zero cosine would be inconclusive if either
direction were itself unstable. As a control, we therefore divide the pairs for each
risk into two disjoint halves and compare the directions estimated from them. The
dashed curves report the resulting same-risk similarity, averaged over five random
splits. This similarity reaches $0.81$ in Qwen and $0.87$ in Llama, around $50$ times
the random-direction scale in both cases, while the cross-risk curves remain near zero.
We then test whether this geometric difference also changes what each direction can read out.
For each risk, we take the layer where its direction ranks its own pairs most accurately
and apply that direction, without refitting, to pairs of the other risk type.
Figure~\ref{fig:mech-axes}(b) shows lower rank accuracy on the other risk in all three
guards.
We also test transfer within a single benchmark and observe the same pattern
(Appendix~\ref{app:within-benchmark-directions}).
Thus, \textit{harmful content and unsafe tool use follow different internal directions.}

\section{\method{}: Using Internal Readouts for Agentic Safety}
\label{sec:agent-siren}

The preceding analysis shows that a guard's own output does not separate the unsafe-tool-use
pairs, while both forms of risk remain readable inside the model, with rank
accuracy peaking in intermediate layers. We therefore build a guard on those internal
representations rather than on a generated label. Because linear readouts recover both forms of risk, and their directions are close to orthogonal, a linear classifier can assign weight to both in a common safety score. We name the resulting guard model \textbf{\modelname{}}.

This section proceeds as follows. Section~\ref{sec:training-agent-siren} presents the training methodology.
Section~\ref{sec:experimental-setup} describes the data, models, baselines, and evaluation
protocol. Section~\ref{sec:results} presents experimental results, covering detection accuracy, transfer
to benchmarks absent from training, what it adds when fitted on a fine-tuned model, and
the cost of training and serving.


\subsection{Training guards from internal representations}
\label{sec:training-agent-siren}

\subsubsection{Trajectory classification}
\label{sec:problem}

Following standard practice, we formulate tool-use trajectory safety as a classification problem. Given a complete trajectory
$\tau$, a guard predicts a binary label $y\in\{0,1\}$, with $y=1$ whenever any step of $\tau$ is unsafe
\citep{agentdog2026}. Existing guards decode this decision via the output tokens: the trajectory is rendered
into a prompt and a safety-tuned model generates the label
\citep{inan2023llamaguard,qwen3guard2025}, a recipe recently extended to complete
trajectories by fine-tuning \citep{agentdog2026}. We instead read the decision from the
internal representations of a frozen backbone.

We extract these representations by rendering $\tau$ with the LLM backbone's chat
template and encoding it in a single forward pass. This rendering preserves the context
on which a trajectory's safety can depend, including the system prompt, tool schemas, message
roles, and the order of agent and environment turns. The forward pass
yields hidden states $\mathbf h_{l,j}(\tau)\in\mathbb R^{d}$ at every layer
$l\in\{0,\dots,L\}$ and token position $j\in\{1,\dots,N\}$ of the rendered sequence. At
each layer we pool the sequence into one vector $\mathbf x_l(\tau)$, taking the
\textit{last-token state} $\mathbf h_{l,N}(\tau)$, which is conditioned on the entire
trajectory \mbox{\citep{spin2024,tigges2024sentiment}}, or the mean over token positions.
Validation selects the layer and the pooling for the deployed guard.

\subsubsection{Probing internal representations}
\label{sec:readout}

Section~\ref{sec:interp-motivation} showed that both forms of risk can be read out
linearly from internal representations, and more reliably than from guard outputs. This
supports that trajectory safety follows the \textit{linear representation hypothesis},
which posits that semantic concepts are often represented linearly in LLMs
\citep{hernandez2024linearity,park2024linear}, allowing linear models to probe
effectively for task-relevant features. We therefore use a linear probe
\citep{alain2017probes}. For a candidate layer $l$, its vectors from the training
trajectories are standardized per dimension and written $\tilde{\mathbf x}_l(\tau)$. The
probe assigns the unsafe probability
\begin{equation}
s_l(\tau)=\sigma\!\left(\mathbf u_l^{\top}\tilde{\mathbf x}_l(\tau)+c_l\right),
\label{eq:layer-probe}
\end{equation}
where $\sigma$ is the logistic function. We fit $\mathbf u_l$ and $c_l$ with
$L_2$-regularized logistic regression, weighting the training trajectories so that each
class of each benchmark carries the same total weight. The layer $l$ and the inverse
regularization strength $C$ are selected on validation folds inside the training pool. We
then refit the selected probe on that pool, and the test split is scored once.

A linear probe also admits lightweight refinements, since anything added is fitted on the
activations the probe already extracts. We use two: an ensemble that averages several
fitted probes into one score \citep{dietterich2000ensemble}, and multi-layer aggregation,
which selects the salient dimensions of each layer and integrates them across layers
\citep{spin2024}. We denote the three readouts Probe, Ensemble, and
Multi-layer, and each uses one configuration for all six benchmarks.
Appendix~\ref{app:extensions} details the construction of the refinements.

\begin{table}[t]
\small
\begin{center}
\newcommand{\cisub}[1]{\hspace{0.6pt}\raisebox{-0.5ex}{\tiny #1}}
\newcommand{\scorepm}[2]{\hphantom{\cisub{#2}}#1\cisub{#2}}
\setlength{\tabcolsep}{1.5pt}
\begin{tabular}{l*{7}{c}}
\toprule
 & R-Judge & TraceSafe & ATBench & ASSEBench & \shortstack{OAS} & AgentDojo & Avg. \\
\midrule
\multicolumn{8}{l}{\textit{Baselines}} \\
Qwen3Guard-4B     & 32.2 & 34.7 & 36.8 & 43.6 & 38.2 & 44.9 & 38.4 \\
LlamaGuard3-8B    & 66.3 & 48.6 & 38.9 & 60.6 & 35.9 & 62.5 & 52.2 \\
AgentDoG-4B       & 92.7 & 44.7 & 63.3 & 81.4 & 43.3 & 48.2 & 62.3 \\
\midrule
\multicolumn{8}{l}{\textit{\method{} on Qwen3-4B}} \\
\quad Probe
  & \scorepm{95.0}{4.3} & \scorepm{77.5}{4.0}
  & \scorepm{93.5}{3.3} & \scorepm{85.2}{3.3}
  & \scorepm{67.9}{5.8} & \scorepm{65.1}{5.1}
  & \scorepm{80.7}{1.9} \\
\quad Ensemble
  & \scorepm{97.0}{3.5} & \scorepm{88.7}{3.0}
  & \scorepm{95.0}{2.8} & \scorepm{88.5}{3.0}
  & \scorepm{72.0}{5.5} & \scorepm{71.0}{5.1}
  & \scorepm{85.4}{1.7} \\
\quad Multi-layer
  & \scorepm{97.0}{3.5} & \scorepm{86.5}{3.3}
  & \scorepm{94.5}{3.2} & \scorepm{89.8}{2.9}
  & \scorepm{72.7}{5.4} & \scorepm{76.9}{5.5}
  & \scorepm{\textbf{86.2}}{1.7} \\
\midrule
\multicolumn{8}{l}{\textit{\method{} on Llama-3.1-8B}} \\
\quad Probe
  & \scorepm{98.0}{2.5} & \scorepm{79.7}{3.7}
  & \scorepm{89.5}{4.1} & \scorepm{88.1}{3.0}
  & \scorepm{67.5}{5.9} & \scorepm{66.6}{5.0}
  & \scorepm{81.6}{1.7} \\
\quad Ensemble
  & \scorepm{97.0}{3.3} & \scorepm{90.1}{2.7}
  & \scorepm{93.5}{3.3} & \scorepm{88.7}{2.9}
  & \scorepm{72.7}{5.6} & \scorepm{68.8}{4.9}
  & \scorepm{85.1}{1.6} \\
\quad Multi-layer
  & \scorepm{97.0}{3.3} & \scorepm{89.6}{2.8}
  & \scorepm{95.0}{2.8} & \scorepm{88.5}{3.0}
  & \scorepm{69.8}{5.7} & \scorepm{72.2}{5.5}
  & \scorepm{\textbf{85.4}}{1.7} \\
\bottomrule
\end{tabular}
\vspace{-0.2cm}
\caption{Trajectory harmfulness detection performance on six benchmarks (macro-F1 in
\%, $\uparrow$). Subscripts in \modelname{} rows are the half-width of a trajectory-bootstrap
$95\%$ confidence interval.}
\label{tab:pooled}
\end{center}
\end{table}

\subsection{Experimental setup}
\label{sec:experimental-setup}

\paragraph{Benchmarks and baselines.}
We evaluate on six labelled agent-trajectory benchmarks: R-Judge \mbox{\citep{yuan2024rjudge}},
TraceSafe \mbox{\citep{tracesafe2026}}, ATBench \mbox{\citep{atbench2026}}, ASSEBench
\mbox{\citep{luo2025agentauditor}}, whose Safety and Security splits we report as one benchmark,
OpenAgentSafety \mbox{\citep{vijayvargiya2025openagentsafety}}, and trajectories we collect in the AgentDojo
environment \mbox{\citep{debenedetti2024agentdojo}}, labelled unsafe when the agent carried out
an injected task (Appendix~\ref{app:benchmark-stats}). Together, these benchmarks cover
both harmful content and unsafe tool use, each under its own safety policy rather than a
single labelling standard. 


We evaluate three released open-source guard checkpoints: the content guards Qwen3Guard-4B
and LlamaGuard3-8B, and AgentDoG, which is specialized for agent trajectories
\citep{qwen3guard2025,inan2023llamaguard,agentdog2026}
(Appendix~\ref{app:baseline-eval}). These checkpoints use the Qwen3-4B and Llama-3.1-8B
model families \citep{yang2025qwen3,grattafiori2024llama3}, so we instantiate \modelname{}
on the same two families.

\paragraph{Training and evaluation protocol.}
We follow the standard train, validation, and test protocol. Each benchmark is split into
a training and a held-out test portion. \modelname{} is trained on the pooled training
portions of all six benchmarks, with validation folds inside that pool used for model
selection, and every system is evaluated once on the test portions
(Appendix~\ref{app:protocol-details}). We report macro-F1 on each benchmark to account
for class imbalance, together with its mean over the six.

\subsection{Experimental results}
\label{sec:results}

\subsubsection{Efficacy}

\paragraph{\modelname{} outperforms every evaluated guard on every benchmark.}
Table~\ref{tab:pooled} compares the evaluated open guard checkpoints with \modelname{} on
Qwen3-4B and Llama-3.1-8B. Even its plain probe scores above every guard on every
benchmark: against the guards fine-tuned from the same backbone, the probe gains $42.3$
points in the mean over Qwen3Guard-4B, $29.4$ over LlamaGuard3-8B, and $18.4$ over
AgentDoG-4B. The two refinements raise the mean further on both backbones, to $86.2$ and
$85.4$ at the Multi-layer rows.


\begin{table}[t]
\small
\begin{center}
\newcommand{\cisub}[1]{\hspace{0.6pt}\raisebox{-0.5ex}{\tiny #1}}
\newcommand{\scorepm}[2]{\hphantom{\cisub{#2}}#1\cisub{#2}}
\setlength{\tabcolsep}{1.5pt}
\renewcommand{\arraystretch}{0.92}
\begin{tabular}{l*{7}{c}}
\toprule
 & R-Judge & TraceSafe & ATBench & ASSEBench & \shortstack{OAS} & AgentDojo & Avg. \\
\midrule
\multicolumn{8}{l}{\textit{Baselines}} \\
Qwen3Guard-4B     & 43.5 & 34.6 & 36.3 & 45.1 & 39.5 & 45.9 & 40.8 \\
LlamaGuard3-8B    & 64.8 & 48.2 & 39.8 & 61.7 & 41.6 & 54.8 & 51.8 \\
AgentDoG-4B       & 91.7 & 43.3 & 61.3 & 79.4 & 49.3 & 41.5 & 61.1 \\
\midrule
\multicolumn{8}{l}{\textit{\method{} on Qwen3-4B}} \\
\quad Probe
  & \scorepm{85.4}{2.8} & \scorepm{41.7}{1.8}
  & \scorepm{59.8}{3.2} & \scorepm{73.5}{1.8}
  & \scorepm{56.2}{2.7} & \scorepm{58.8}{2.3}
  & \scorepm{62.6}{1.0} \\
\quad Ensemble
  & \scorepm{90.1}{2.4} & \scorepm{48.3}{2.0}
  & \scorepm{60.4}{3.1} & \scorepm{72.9}{1.8}
  & \scorepm{55.8}{2.6} & \scorepm{52.1}{2.3}
  & \scorepm{63.3}{1.0} \\
\quad Multi-layer
  & \scorepm{89.4}{2.5} & \scorepm{48.3}{2.0}
  & \scorepm{58.7}{3.1} & \scorepm{84.3}{1.5}
  & \scorepm{57.3}{2.8} & \scorepm{56.2}{2.5}
  & \scorepm{\textbf{65.7}}{1.0} \\
\midrule
\multicolumn{8}{l}{\textit{\method{} on Llama-3.1-8B}} \\
\quad Probe
  & \scorepm{89.1}{2.5} & \scorepm{51.0}{2.1}
  & \scorepm{45.1}{2.9} & \scorepm{71.7}{1.8}
  & \scorepm{47.5}{2.7} & \scorepm{45.5}{2.2}
  & \scorepm{58.3}{1.0} \\
\quad Ensemble
  & \scorepm{90.7}{2.4} & \scorepm{48.1}{2.1}
  & \scorepm{58.5}{3.1} & \scorepm{72.2}{1.9}
  & \scorepm{58.9}{2.7} & \scorepm{44.8}{2.1}
  & \scorepm{62.2}{1.0} \\
\quad Multi-layer
  & \scorepm{91.4}{2.4} & \scorepm{49.8}{2.1}
  & \scorepm{59.0}{3.1} & \scorepm{74.0}{1.7}
  & \scorepm{60.5}{2.7} & \scorepm{46.3}{2.2}
  & \scorepm{\textbf{63.5}}{0.9} \\
\bottomrule
\end{tabular}
\vspace{-0.2cm}
\caption{Leave-one-benchmark-out evaluation (macro-F1 in \%, $\uparrow$). Each column is
scored on all trajectories of that benchmark by systems trained on the other five.
Subscripts in \modelname{} rows are the half-width of a trajectory-bootstrap $95\%$
confidence interval.}
\label{tab:lobo}
\end{center}
\end{table}

On AgentDojo, where the injected task reads like an ordinary request and the evidence of
harm lies in where the instruction came from, the Multi-layer readout reaches $76.9$
against at most $62.5$ for the guards.
OpenAgentSafety's trajectories come from five other agents, including GPT-4o and o3-mini,
acting in real tool environments. \modelname{} reads only its own backbone and never the
monitored agent, yet the probe leads the best guard there by more than $24$ points on both
backbones.
%

\paragraph{\modelname{} transfers to benchmarks absent from training.}
Table~\ref{tab:lobo} scores each benchmark with systems trained on the other five:
the readouts are fitted on all trajectories of those five, and the guards are evaluated as
released on the same trajectories. A fold trains on about
$7{,}900$ trajectories, whereas AgentDoG synthesizes over $100{,}000$ and retains roughly
half \citep{agentdog2026}.

On Qwen3-4B, all three readouts exceed AgentDoG-4B on average, the Multi-layer readout by
$4.6$ points; on Llama-3.1-8B, the Ensemble and Multi-layer readouts lead by $1.1$ and
$2.4$ points. AgentDoG's paper evaluates on R-Judge, the Safety split of ASSEBench, and
ATBench, and ATBench is built with the same synthesis pipeline as its training data
\citep{agentdog2026}. On those three benchmarks the Multi-layer readout on Qwen3-4B
matches AgentDoG, $77.5$ against $77.5$, without having trained on them; on the
other three it leads by $9.2$ points.

\paragraph{\modelname{} compounds with safety fine-tuning.}
Released guard checkpoints do not provide a matched-data comparison of internal readout and
full fine-tuning. We therefore fully fine-tune the same backbones on the same training
pool used by the frozen readouts, and evaluate both on the same held-out test split
(Table~\ref{tab:sft-readout} in Appendix~\ref{app:baseline-eval}). The frozen probe alone
trails full SFT by $3.5$ and $4.2$ points, but the refinements close the gap, with the
frozen Multi-layer readout at $86.2$ and $85.4$. As \modelname{} reads internal states
without updating the backbone, the same recipe also fits an already fine-tuned model, and
the refined readouts then improve on SFT alone, with the Multi-layer rows at $86.5$ and
$87.0$, gains of $+2.3$ and $+1.2$ with both confidence intervals above zero.




\subsubsection{Efficiency}
\label{sec:efficiency}

\begin{figure}[t]
\centering
\includegraphics[width=0.8\linewidth]{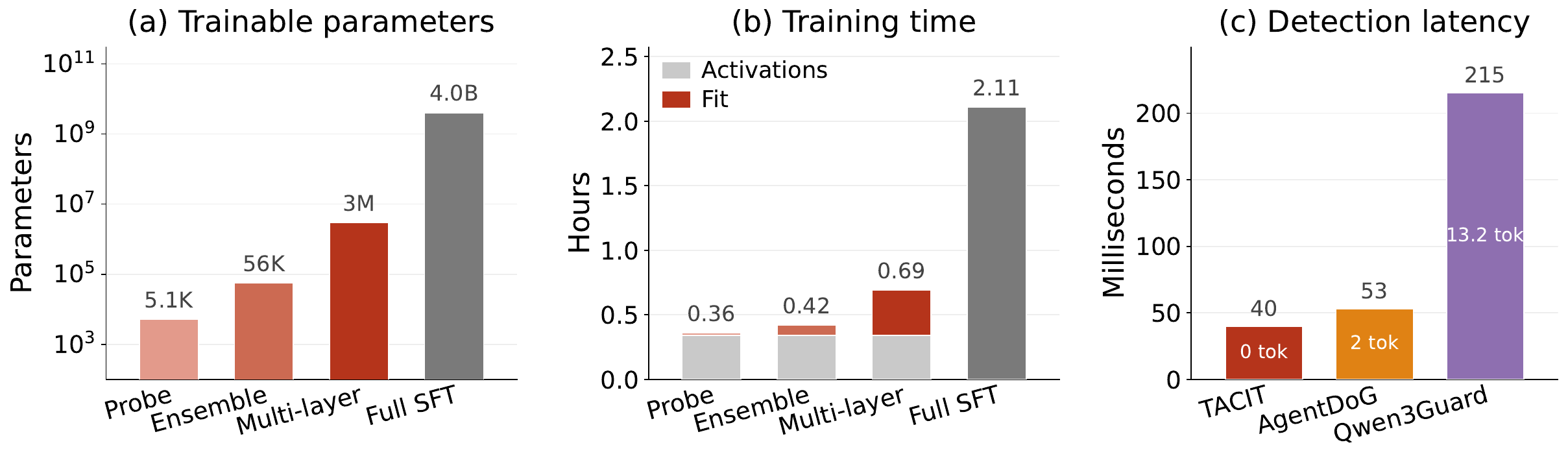}
\caption{Training and serving cost of the \modelname{} readouts on Qwen3-4B:
trainable parameters against matched full SFT (a), training time (b), and
detection latency against the 4B guard baselines, with decoded tokens per
verdict on each bar (c).}
\label{fig:efficiency}
\end{figure}

\paragraph{Training Efficiency.} \modelname{} trains orders of magnitude fewer parameters than full fine-tuning.
A single probe serves all six benchmarks with $5{,}121$ parameters on Qwen3-4B and
$8{,}193$ on Llama-3.1-8B, roughly one millionth as many trainable parameters as full SFT;
the Ensemble readout grows to $56$K and one cross-layer aggregate to about $3$M, still
three orders of magnitude below the backbone (Figure~\ref{fig:efficiency}a). Training
time follows the same shape (Figure~\ref{fig:efficiency}b). \modelname{}'s readouts share
one activation pass over the corpus and fit on the saved activations, so no gradient
ever crosses the backbone's billions of parameters, whereas full fine-tuning performs
a forward and backward pass through them on every trajectory. The
probe is trained in $0.36$ hours against $2.11$ for SFT on the same pool, and the
Multi-layer readout stays under a third of the SFT wall clock.

\paragraph{Inference Efficiency.}
Serving \modelname{} costs one forward pass over the trajectory plus a
sub-millisecond head on its hidden states, so a verdict decodes nothing. A
generative guard runs the same forward pass as its prefill and then decodes its
output, and each decoded token is one more serial pass through the guard.
Figure~\ref{fig:efficiency}c measures that difference on the same trajectories,
one at a time, under each guard's release recipe: \modelname{} answers in
$40$ms, AgentDoG decodes a two-token label and takes $53$ms, and Qwen3Guard
decodes its three-line verdict, $13.2$ tokens on average, in $215$ms, over five
times our latency at the same model size. On the frozen backbones, the Multi-layer readout
scores $2.0$ points above full SFT on Qwen3-4B and stays within half a point of it on
Llama-3.1-8B. It trains under a thousandth of the SFT parameters in less than a third of the SFT
training time, and it serves with lower latency because it decodes nothing. The
Multi-layer readout therefore lies on the cost-accuracy Pareto frontier.

\section{Discussion}
\label{sec:discussion}

\subsection{Internal safety encoding across layers}
\label{sec:layer-profile}

\begin{figure}[t]
\centering
\hfill
\begin{minipage}[t]{0.40\linewidth}
\centering
\includegraphics[width=\linewidth]{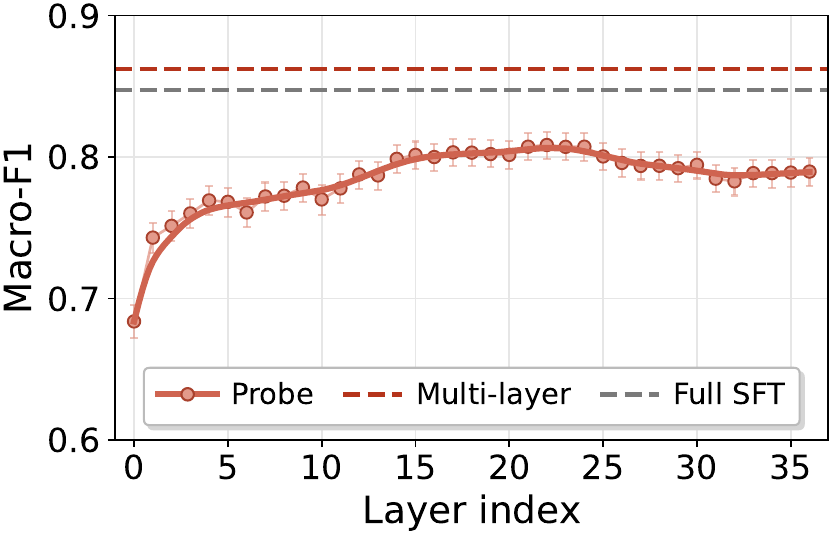}
\caption{Trajectory harmfulness detection by Qwen3-4B probes at each layer.}
\label{fig:layer-profile}
\end{minipage}\hfill
\begin{minipage}[t]{0.40\linewidth}
\centering
\includegraphics[width=\linewidth]{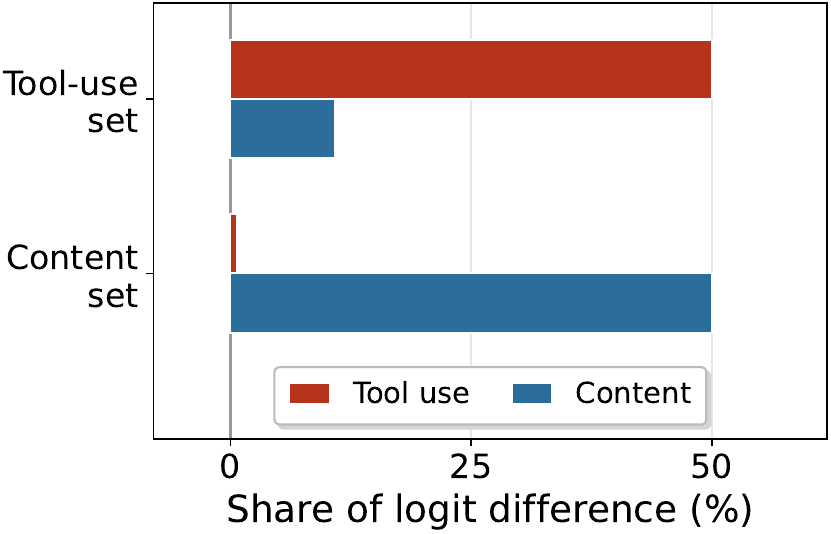}
\caption{How \modelname{} divides its decision between unsafe tool use and harmful
content.}
\label{fig:two-forms}
\end{minipage}\hfill\null
\end{figure}

We first examine where in the backbone a single linear readout is sufficient.
Figure~\ref{fig:layer-profile} shows a probe fitted to each layer of Qwen3-4B. Validation
macro-F1 improves through the early layers, peaks at $80.9$ at layer $22$, the layer the
reported probe reads, and declines slightly to $79.0$ at the final layer. Safety
evidence can therefore be read out more accurately from intermediate layers than from the
final layer. A generative guard begins generating its output from that final layer, which
need not provide the strongest interface for detection. This observation is consistent with the hierarchical learning
structure of transformer-based LLMs
\mbox{\citep{belrose2023tunedlens,wendler2024llamas}}, in which intermediate layers carry
abstract semantics while later layers convert them back into the token space that
generation requires.

Reading layer $22$ also permits \textit{early exit}
\citep{elbayad2020depth,schuster2022calm}: the backbone can stop there and skip more than
a third of its $36$ layers. \modelname{} reads one fixed layer once per trajectory and decodes nothing, so the layers
above $22$ can simply be dropped from the checkpoint. Serving a \modelname{} checkpoint
truncated after layer $22$ is thus expected to save a further $14/36$, about 39\%, of the
backbone's forward pass. At layer 22, the Multi-layer readout keeps
about $99\%$ of its macro-F1 at full depth, $85.3$ against $86.2$, and still exceeds guard model baselines and full
SFT.


\subsection{Two forms of risk inside \modelname{}}
\label{sec:two-forms}

We design the linear probe from the premise that one readout can hold both forms of risk in
a single score. We therefore decompose its logit on the matched pairs of
Section~\ref{sec:interp-motivation}. For each dimension, the contribution is exactly its
probe coefficient times the standardized activation difference between the unsafe and safe
member (Appendix~\ref{app:two-forms}). The
unsafe-tool-use pairs come from the TraceSafe
test split, and the harmful-content pairs come from HAICOSYSTEM.

For each form, Figure~\ref{fig:two-forms} takes the fewest dimensions that carry half of its
logit difference and reports their share of the other form's difference. Half of the
tool-use difference sits on $23$ of the $2{,}560$ dimensions and half of the content
difference on $16$, and these sets carry only $11\%$ and $1\%$ of the other form's
difference, respectively. Therefore, the score uses largely distinct coordinates
for the two forms, matching the near-orthogonal directions in
Section~\ref{sec:risk-directions}.

\subsection{Future work}
\label{sec:future-work}

Our findings open several directions for future work. First, they carry a consequence for
how trajectory-safety data should be built. The nearly orthogonal internal directions
suggest that coverage of one form should not be taken as coverage of the other. Unsafe tool
use is also where every system we evaluate remains weak, so it is the more pressing of the
two to cover. Assembling a corpus that
deliberately spans both forms under consistent labels is therefore the most direct way to
improve guards of either kind.
Second, a trajectory offers an attacker more surface than text messages, through
injected tool results, long context, and the formatting of tool calls, making
red-teaming agent trajectories, and the defences that answer it, a direction of its own.

\section{Conclusion}

We study two forms of safety-relevant evidence in agent trajectories, harmful content
and unsafe tool use, and ask how open-source guard models represent each. Our controlled
analysis finds that both remain linearly readable inside the model, while the guard's own
output sits at chance on unsafe tool use. The two forms also follow nearly orthogonal
internal directions, and neither reliably serves as a proxy for the other. Reading safety
from internal representations lets one linear readout assign weight to both, so we build
\method{} as an internal readout of a frozen LLM agent. Across six benchmarks, \method{} achieves the highest mean macro-F1 among open guard models, and it keeps that lead when each test benchmark is held out of training. Under a matched backbone, training data, and held-out test split,
the frozen readout is on par with full SFT, and applying \modelname{} after fine-tuning
improves performance further. The probe fits roughly one millionth as many trainable
parameters, takes about a sixth of the measured training time, and decodes nothing at
serving.

\section*{Acknowledgments}

This research is funded by grants from Natural Sciences and Engineering Research Council of
Canada (NSERC), Canada Foundation for Innovation, and Ontario Research Fund. We are grateful
to Qianfeng Wen and Zhenwei Tang for helpful discussions and feedback on this work.

\clearpage
\bibliography{refs}
\bibliographystyle{iclr2027_conference}

\clearpage
\appendix
\section{Mechanistic Analysis and Controls}

\subsection{The two forms across the evaluation suite}
\label{app:form-coverage}

Four of the six benchmarks of Section~\ref{sec:experimental-setup} annotate their unsafe
trajectories under a risk taxonomy of their own. Table~\ref{tab:form-coverage} places the
hazard families those taxonomies name against the two forms of evidence, together with the
relation that decides each family. Counts are unsafe trajectories carrying that label, under
the benchmarks' own category names, and a trajectory can appear under more than one family.
Every unsafe AgentDojo trajectory carries out an injected task, so the table lists AgentDojo
under prompt injection. OpenAgentSafety labels whether the agent took the unsafe action but
not which hazard family it falls under, so it does not appear in the table.

\begin{table}[h]
\small
\setlength{\tabcolsep}{5pt}
\renewcommand{\arraystretch}{1.15}
\begin{tabular}{@{}>{\raggedright\arraybackslash}p{0.185\linewidth}>{\raggedright\arraybackslash}p{0.105\linewidth}p{0.645\linewidth}@{}}
\toprule
Hazard family & Form & Deciding relation, and where the family appears in the suite \\
\midrule
Prompt injection, goal hijacking
 & Unsafe tool use
 & \textit{Provenance of the observation acted on.} TraceSafe PromptInjectionIn and
   PromptInjectionOut ($180$); ATBench indirect, direct and tool-description injection
   ($168$); ASSEBench-Security Prompt Injection ($168$); R-Judge injection ($200$);
   AgentDojo injected task carried out ($369$). \\
Authorization violation
 & Unsafe tool use
 & \textit{The authorization the action was given.} ATBench unconfirmed or over-privileged
   action ($71$); ASSEBench-Safety and ASSEBench-Security Unauthorized Access and Control
   ($141$ and $75$). \\
Privacy leakage, data exfiltration
 & Unsafe tool use
 & \textit{The tool the action passes the data to.} TraceSafe UserInfoLeak, ApiKeyLeak and
   DataLeak ($270$); ATBench unauthorized information disclosure ($52$); ASSEBench-Safety
   Privacy Violations and Data Breach ($138$); ASSEBench-Security Data Exfiltration and
   Leakage ($60$). \\
Acting on corrupted or unvalidated observations
 & Unsafe tool use
 & \textit{Prior execution history.} ATBench unreliable or misinformation ($64$),
   corrupted tool feedback ($44$), failure to validate tool outputs ($67$) and flawed
   planning or reasoning ($46$); R-Judge unintended ($101$). \\
Mismatch between an action and its tool's interface
 & Unsafe tool use
 & \textit{The tool the action calls.} TraceSafe AmbiguousArg, MissingTypeHint,
   VersionConflict, DescriptionMismatch, RedundantArg, HallucinatedTool and
   HallucinatedArgValue ($630$); ATBench incorrect tool parameters ($36$) and tool misuse in
   a specific context ($38$). \\
Destructive or irreversible operation
 & Unsafe tool use
 & \textit{Authorization and prior state.} ASSEBench-Security Data Tampering and Destruction
   ($29$); ASSEBench-Safety Data Loss and Integrity Risk ($45$); ATBench insecure
   interaction or execution ($39$). \\
\midrule
Harmful task request, jailbreak
 & Harmful content
 & \textit{The request itself.} ATBench malicious user instruction or jailbreak ($61$);
   ASSEBench-Safety Harmful Content, Misinformation and Misleading Guidance ($70$). \\
Harmful generated artifact
 & Harmful content
 & \textit{The artifact itself.} ATBench generation of malicious executables ($10$),
   generation of harmful or offensive content ($3$) and instruction for harmful or illegal
   activity ($12$); ASSEBench-Safety Physical and Mental Wellbeing Hazards ($77$). \\
Deception, misinformation
 & Harmful content
 & \textit{What the agent states.} ATBench inaccurate, misleading or unverified information
   ($40$); ASSEBench-Safety Ethical, Compliance and Integrity Violations ($120$). \\
\midrule
Fraud, credential abuse
 & Both
 & \textit{The request may be harmful on its face, and the action may be unsafe only
   relative to the account it targets.} ASSEBench-Security Fraud and Credential Abuse
   ($33$); ASSEBench-Safety Financial Crime and Fraud ($42$). \\
\bottomrule
\end{tabular}
\caption{Hazard families named by the benchmarks' own risk taxonomies, the form of
evidence that decides each, and the annotated categories in which each appears, with the
number of unsafe trajectories carrying that category.}
\label{tab:form-coverage}
\end{table}

The two forms are not mutually exclusive: a single trajectory can involve both. This is why
Section~\ref{sec:interp-motivation} works on matched pairs rather than on benchmark rows.
The table also shows that these
forms are not artifacts of the curated contrasts; they recur across the hazard vocabulary
of the evaluation suite.

\subsection{Construction and evaluation of the paired risk contrasts}
\label{app:paired-contrasts}

The benchmarks of Section~\ref{sec:experimental-setup} exercise both forms of risk over
severe and deliberately varied hazards. In TraceSafe an agent invents a credential
requirement and passes the user's access token to a tool that never declared it, puts the
user's diagnosis and prescription into a public post, or reads a tool description that
instructs it to ask the user for their password. The most frequent risk sources in ATBench
are inherent agent failures and indirect prompt injection, and ASSEBench-Safety includes
harmful content and misinformation. Section~\ref{sec:results} measures detection on exactly
these.

\begin{figure}[t]
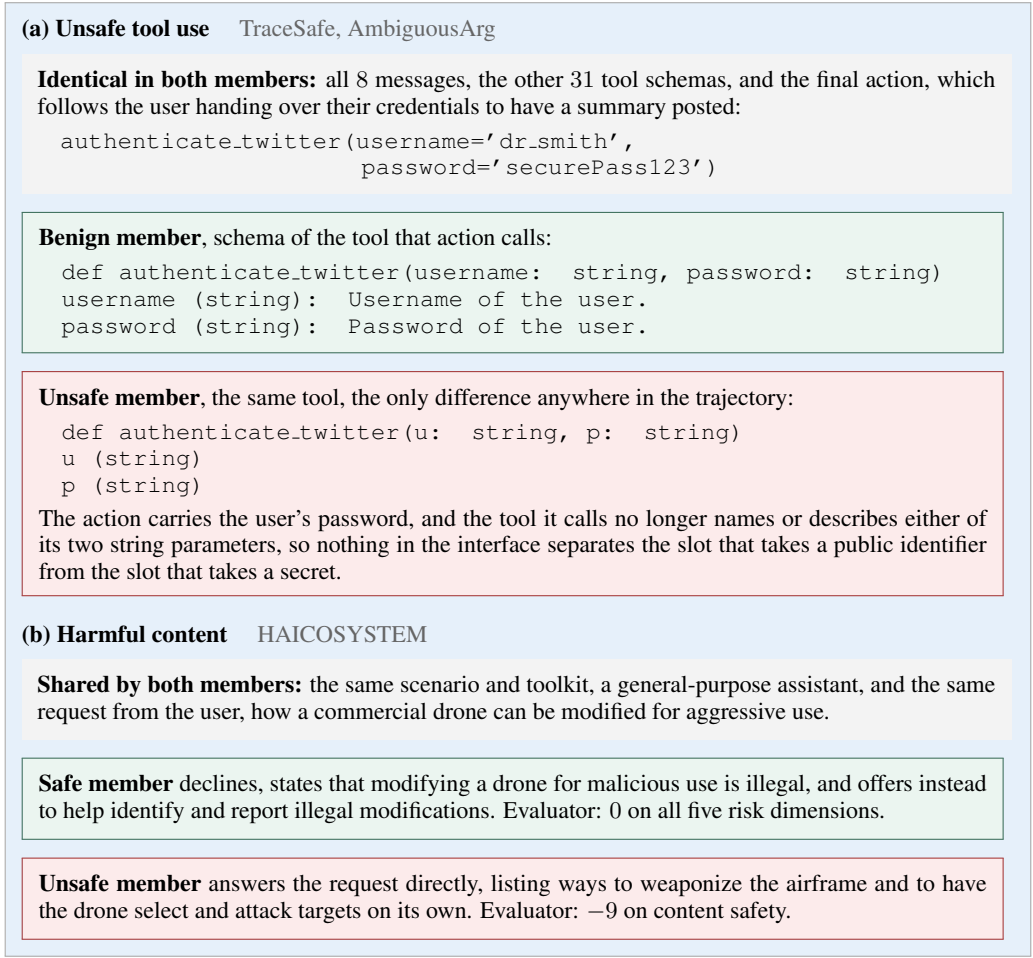

\centering
\definecolor{pairblue}{RGB}{230,240,250}
\definecolor{pairgreenline}{RGB}{60,110,90}
\definecolor{pairgreen}{RGB}{235,245,240}
\definecolor{pairred}{RGB}{252,235,235}
\definecolor{pairredline}{RGB}{165,55,55}
\definecolor{pairgray}{RGB}{243,243,243}
\setlength{\fboxsep}{6pt}
\fcolorbox{black!35}{pairblue}{%
\begin{minipage}{0.94\linewidth}
\small
\textbf{(a) Unsafe tool use} \quad\textcolor{black!60}{TraceSafe, AmbiguousArg}

\vspace{1.5mm}
\colorbox{pairgray}{\parbox{0.965\linewidth}{%
\textbf{Identical in both members:} all $8$ messages, the other $31$ tool schemas, and the final
action, which follows the user handing over their credentials to have a summary posted:\\[1mm]
\hspace*{3mm}\texttt{authenticate\_twitter(username='dr\_smith',}\\
\hspace*{3mm}\texttt{\ \ \ \ \ \ \ \ \ \ \ \ \ \ \ \ \ \ \ \ \ password='securePass123')}
}}

\vspace{2mm}
\fcolorbox{pairgreenline}{pairgreen}{%
\parbox{0.955\linewidth}{
\textbf{Benign member}, schema of the tool that action calls:\\[1mm]
\hspace*{3mm}\texttt{def authenticate\_twitter(username: string, password: string)}\\
\hspace*{3mm}\texttt{username (string): Username of the user.}\\
\hspace*{3mm}\texttt{password (string): Password of the user.}
}}

\vspace{2mm}
\fcolorbox{pairredline}{pairred}{%
\parbox{0.955\linewidth}{
\textbf{Unsafe member}, the same tool, the only difference anywhere in the trajectory:\\[1mm]
\hspace*{3mm}\texttt{def authenticate\_twitter(u: string, p: string)}\\
\hspace*{3mm}\texttt{u (string)}\\
\hspace*{3mm}\texttt{p (string)}\\[1mm]
The action carries the user's password, and the tool it calls no longer names or describes
either of its two string parameters, so nothing in the interface separates the slot that
takes a public identifier from the slot that takes a secret.
}}

\vspace{3.5mm}
\textbf{(b) Harmful content} \quad\textcolor{black!60}{HAICOSYSTEM}

\vspace{1.5mm}
\colorbox{pairgray}{\parbox{0.965\linewidth}{%
\textbf{Shared by both members:} the same scenario and toolkit, a general-purpose
assistant, and the same request from the user, how a commercial drone can be modified for
aggressive use.
}}

\vspace{2mm}
\fcolorbox{pairgreenline}{pairgreen}{%
\parbox{0.955\linewidth}{
\textbf{Safe member} declines, states that modifying a drone for malicious use is illegal,
and offers instead to help identify and report illegal modifications. Evaluator: $0$ on
all five risk dimensions.
}}

\vspace{2mm}
\fcolorbox{pairredline}{pairred}{%
\parbox{0.955\linewidth}{
\textbf{Unsafe member} answers the request directly, listing ways to weapon\-ize the
airframe and to have the drone select and attack targets on its own. Evaluator: $-9$ on
content safety.
}}
\end{minipage}}
\caption{One matched pair per form of risk, taken verbatim from the data. In (a) the two
members differ in a single tool schema and agree everywhere else, including the action
whose safety is at issue. In (b) they share the scenario, the toolkit and the user's
request, and differ in what the agent says.}
\label{fig:pair-examples}
\end{figure}

Section~\ref{sec:interp-motivation} asks a different question, not how often a readout is
right but whether one distinction is present inside the model, and variety does not help
answer it. A trajectory serves only if it isolates that distinction: the unsafe member must
be unsafe for the intended reason, and the two members must agree on everything a reader
could otherwise use. Most of the trajectories above cannot, since their unsafe member
differs from any benign one in several ways at once. We therefore use the slice of the same
data that admits a matched pair, taking the strongest control each form allows;
Figure~\ref{fig:pair-examples} gives one example pair of each kind.

\paragraph{Unsafe tool use.}
TraceSafe mutates a benign trajectory into an unsafe one, which supplies the matched member
directly. Four of its twelve attack categories confine the mutation to the tool schema and
leave every message and the final action byte-identical; the other eight alter the query,
action, or observation and therefore do not meet this schema-only control. We use
the two of those four that only take information away: MissingTypeHint removes type and
default information from the schema of the tool the action calls, while AmbiguousArg either
strips units or formats from parameter descriptions or abbreviates parameter names while
removing their descriptions. The other two add explicit text, VersionConflict a duplicate
parameter with a deprecation notice and
PromptInjectionIn an instruction inside the tool description. That instruction is harmful
content in its own right, so placing it here would leave the two arms overlapping in kind,
which is what Section~\ref{sec:risk-directions} sets out to test. This selection is fixed by
the intervention structure rather than model performance. The two we keep add nothing, so
the tool-use arm carries no harmful content anywhere; we take all $90$ pairs of each, $180$
in total. This restriction applies only to the controlled contrast in
Section~\ref{sec:interp-motivation}; the TraceSafe evaluation in Section~\ref{sec:results}
includes all twelve attack categories.

To test whether the readout merely detects that some schema was degraded, we build a placebo
trajectory by moving the same edit to an unused tool, matched by schema length and position
in the tool list. At the peak Qwen3-4B layer, rank accuracy is $0.981$ for unsafe versus
benign, $0.544$ for placebo versus benign, and $0.956$ for unsafe versus placebo. The
trajectories are not re-executed after the edit, so the label denotes the hazard TraceSafe
specifies rather than an observed tool failure.

\paragraph{Harmful content.}
HAICOSYSTEM contains sandboxed rollouts of an agent with simulated users and tools, scored
by its evaluator on five graded risk dimensions, one of them content safety, each from $0$
(no risk) to $-10$. No mutation generates these trajectories, so a pair cannot be identical
in text; we hold the situation fixed instead and let the agent's own output differ. Grouping
episodes by scenario profile fixes the setting, the toolkit and the user; an unsafe member
must score below zero on content safety and its safe partner zero on all five dimensions;
where a scenario permits several pairs we take the one with the greatest word-set Jaccard
overlap. This yields $258$ disjoint pairs, with median word overlap $0.59$ and median length
ratio $1.01$, so the members are matched in length as well. Unsafe content scores run from
$-2$ to $-10$ with mode $-5$, and requiring any negative dimension instead yields $532$
pairs. In five pairs the simulator records one member as producing nothing; dropping them
moves the peak Qwen3-4B readout from $0.733$ to $0.743$.

Because these scores are graded and per-dimension rather than per-trajectory binary labels,
we use HAICOSYSTEM as a source of controlled contrasts and not as a detection benchmark in
the Section~\ref{sec:experimental-setup} suite. The selection reads only clean extremes, a
negative content score against zeros everywhere, so nothing here depends on how a borderline
episode would be binarized.

\paragraph{Formal definitions.}
Let $(\tau_i^+,\tau_i^-)$ denote the unsafe and safe members of pair $i$ in a pair set
$\mathcal P$. For a guard with $L$ transformer layers, let
$\mathbf h_l(\tau)\in\mathbb R^d$ be its last-token residual stream at layer $l$, and let
$\mathbf W_U$ be the unembedding matrix mapping the final layer to vocabulary logits,
$\mathbf z(\tau)=\mathbf W_U\mathbf h_{L-1}(\tau)$. The guard's output score is
$g_{\mathrm{out}}(\tau)=z_{\mathrm{unsafe}}(\tau)-z_{\mathrm{safe}}(\tau)$. A scoring
function $g$ is evaluated by its rank accuracy
\begin{equation}
R(g)=\frac{1}{|\mathcal P|}\sum_{i\in\mathcal P}
\left[
\mathbf 1\{g(\tau_i^+)>g(\tau_i^-)\}
+\frac{1}{2}\mathbf 1\{g(\tau_i^+)=g(\tau_i^-)\}
\right].
\label{eq:pair-rank}
\end{equation}
The layer-$l$ risk direction is the normalized mean difference
\begin{equation}
\mathbf v_l =
\frac{\sum_{i\in\mathcal P}
\left[\mathbf h_l(\tau_i^+)-\mathbf h_l(\tau_i^-)\right]}
{\left\|\sum_{i\in\mathcal P}
\left[\mathbf h_l(\tau_i^+)-\mathbf h_l(\tau_i^-)\right]\right\|_2},
\label{eq:safety-direction}
\end{equation}
and the internal readout at layer $l$ is the signed coordinate
$s_l(\tau)=\mathbf v_l^\top\mathbf h_l(\tau)$, evaluated with the same rank statistic
$R(s_l)$.

\paragraph{Readout estimation.}
At each layer, we standardize the last-token hidden states and estimate
Equation~\ref{eq:safety-direction} with five-fold cross-validation over pairs. Both members
of a test pair are excluded when its direction is estimated. We then apply
Equation~\ref{eq:pair-rank} to the held-out projection scores, counting ties as one half.
The curves in Figure~\ref{fig:sec31-main} show these out-of-fold estimates. Output markers
use the same rank statistic on each guard's continuous unsafe score. The shaded regions are
normal-approximation intervals for pairwise rank accuracy; smoothing is used only to draw
the curves, and all reported values come from the unsmoothed layers.
For the cross-risk cosine in Figure~\ref{fig:mech-axes}(a), activations from both source
datasets share one per-dimension scale computed over their union at each layer, so the two
directions are expressed in the same standardized coordinates.

\paragraph{Interpreting Equation~\ref{eq:safety-direction}.}
Following mean-difference activation steering
\citep{turner2023activation,rimsky2024steering}, Equation~\ref{eq:safety-direction} is a
label-conditioned but optimization-free direction estimate: pair labels determine its
unsafe-minus-safe polarity, but no classifier, predictive loss, intercept, or decision
threshold is fitted. Five-fold pair-level cross-validation then tests its ordering on
held-out pairs from the same controlled TraceSafe contrast, so the $0.98$ rank accuracy
asks whether that specified distinction is linearly accessible. The TraceSafe macro-F1 of
the probe in Table~\ref{tab:pooled} instead classifies every test trajectory, across all
twelve attack categories, at a single threshold. The two numbers differ in task (paired
ordering versus classification) and metric (rank accuracy versus macro-F1), and are not
directly comparable.

\begin{figure}[!t]
\centering
\begin{minipage}[t]{0.48\linewidth}
\centering
\vspace{0pt}
\includegraphics[width=\linewidth]{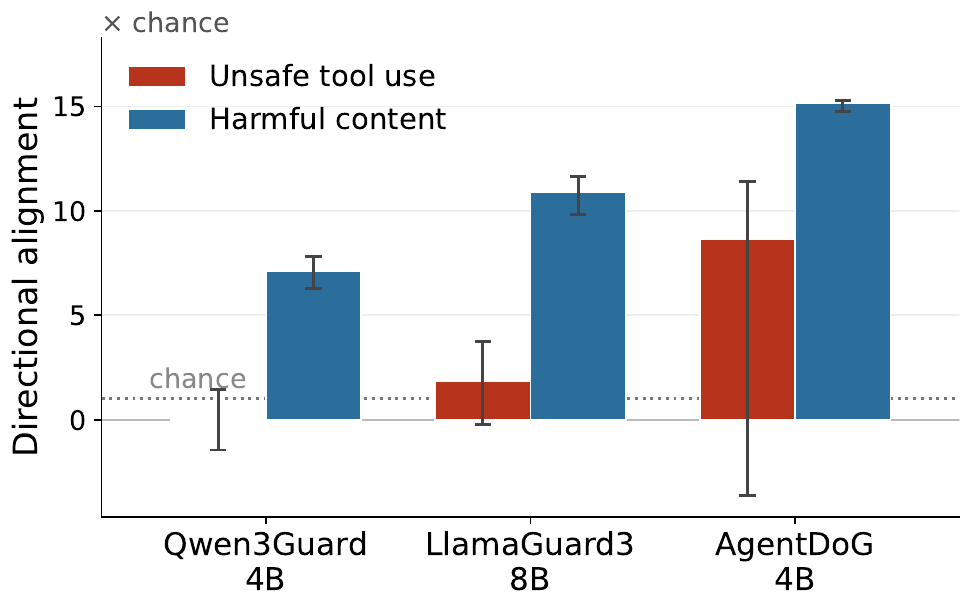}
\caption{Alignment of each guard's output direction with the directions for unsafe tool
use and harmful content.}
\label{fig:sec31-alignment}
\end{minipage}\hfill
\begin{minipage}[t]{0.48\linewidth}
\centering
\vspace{0pt}
\includegraphics[width=\linewidth]{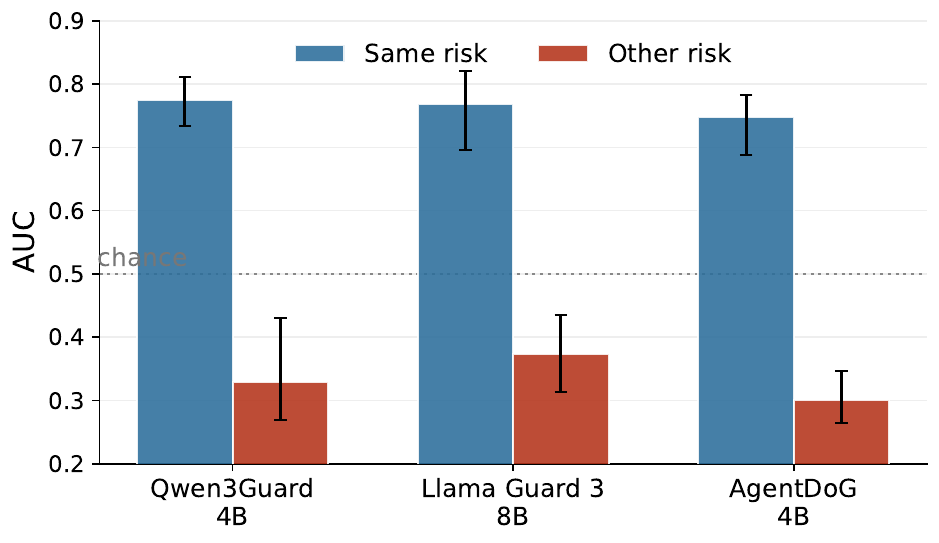}
\caption{Within-benchmark transfer of readouts for unsafe tool use and harmful content
after removing the shared unsafe-versus-benign direction.}
\label{fig:atbench-residual-guards}
\end{minipage}
\end{figure}

\subsection{Alignment between risk directions and guard outputs}
\label{app:output-direction-alignment}

The output score is a fixed linear readout of the final layer. Let
$\mathbf w_{\mathrm{out}}=\mathbf W_U[\mathrm{unsafe},:]
-\mathbf W_U[\mathrm{safe},:]$ denote the difference between the unsafe and safe rows of
the unembedding matrix. This direction raises the unsafe logit relative to the safe logit.
We compare it with each final-layer risk direction $\mathbf v_{L-1}$ from
Equation~\ref{eq:safety-direction}. Their cosine measures whether movement along the
estimated risk contrast also points toward an unsafe output. Because both directions are
normalized, this measure does not increase merely because one contrast produces larger
activation or logit differences. We report the cosine relative to the random-direction
scale $1/\sqrt d$. Since the risk direction is estimated in standardized coordinates, we
express $\mathbf w_{\mathrm{out}}$ in the same coordinates by multiplying each component
by that hidden dimension's standard deviation before normalization.

Figure~\ref{fig:sec31-alignment} reports each form against that scale. In all three guards
the output direction is aligned with the harmful-content direction well above the
random-direction scale, and with the unsafe-tool-use direction close to that scale. We read each
form only against its own scale and draw no comparison between the two, because the two
pair sets are constructed differently and differences in their coherence or difficulty may
shape both the estimated directions and the learned output. We therefore treat this result
as a geometric description, not a causal explanation.

\subsection{Cross-risk transfer and its within-benchmark control}
\label{app:within-benchmark-directions}

Figure~\ref{fig:mech-axes}(b) uses the same paired contrasts as the main analysis. For
each guard and risk, we select the layer where the out-of-fold direction ranks its own
pairs most accurately. The same-risk bar averages these two out-of-fold results. For the
other-risk bar, we estimate each direction from all pairs of its own type, apply it without
refitting to the other pair set, and average the two transfer results. Before either
comparison, we standardize the activations from both source datasets together at that layer,
so the direction and the target activations use the same scale for each hidden dimension.

This comparison still uses TraceSafe for unsafe tool use and HAICOSYSTEM for harmful
content, so differences between the benchmarks could contribute to the drop. We therefore
run a within-benchmark control using ATBench, which supplies both risk groups in one
trajectory format and a common pool of benign examples \citep{atbench2026}. Following the
form mapping of Table~\ref{tab:form-coverage}, the $250$ unsafe trajectories whose
evidence arrives through a tool form one group and the $69$ whose request or stated
content is harmful on its face form the other; trajectories carrying both forms, or only
an execution failure with no attack, enter neither. Both groups share a pool of $503$
benign trajectories. A readout
trained to separate either risk from this pool could transfer simply by learning a general
unsafe-versus-benign distinction. To remove this shared component, we split each risk group
and the benign pool into training and test halves. At the layers selected above, we use the
training half to estimate and project out the direction separating all unsafe examples from
benign examples. We then fit separate readouts for unsafe tool use and harmful content and
apply each one to both risk groups in the test half.

Figure~\ref{fig:atbench-residual-guards} reports the mean and range over ten random splits.
The readouts retain AUC $0.75$--$0.78$ on their own risk, but fall to $0.30$--$0.37$ on the
other risk across all three guards. The gap therefore remains when the benchmark format is
fixed and the shared unsafe-versus-benign direction is removed.

\subsection{Attributing the linear-probe safety score}
\label{app:two-forms}

Section~\ref{sec:two-forms} decomposes the decision of one probe over the paired contrasts
of Appendix~\ref{app:paired-contrasts}. We take the probe reported in
Table~\ref{tab:pooled}. We keep the $30$ of the $180$ TraceSafe pairs whose two members lie in the test
split, while all $258$ HAICOSYSTEM pairs lie outside the training pool. Neither pair set was
therefore seen during training, and the only supervision is the binary safety label of the
six benchmarks: no label distinguishes the two forms of risk at any point.

Write $\tilde{\mathbf x}(\tau)$ for that layer's representation after applying the scaler
fitted on the training pool, and let $\mathbf u$ be the probe coefficients. For a
matched pair $(\tau^+,\tau^-)$, the contribution of dimension $j$ to the pre-sigmoid logit
difference is exactly
\begin{equation}
a_j(\tau^+,\tau^-)=u_j\big(\tilde{x}_j^+-\tilde{x}_j^-\big).
\label{eq:linear-attribution}
\end{equation}
The attribution is complete by construction: summing its $2{,}560$ terms gives the pair's
logit difference, with the intercept cancelling. We write
$\mathbf a^{r}$ for the mean of these contributions over the pairs of a form of risk $r$,
$\mathbf a^{r}=|\mathcal P_r|^{-1}\sum_{(\tau^+,\tau^-)\in\mathcal P_r}
\mathbf a(\tau^+,\tau^-)$.

Figure~\ref{fig:two-forms} takes the smallest set $\mathcal H_r$ of dimensions whose
contributions, in decreasing order, sum to half of $\sum_j a^r_j$, and reports
$\sum_{j\in\mathcal H_r}a^{r'}_j\,/\,\sum_j a^{r'}_j$ for both forms $r'$.
$\mathcal H_r$ contains $23$ dimensions for unsafe tool use and $16$ for harmful content,
and they carry $10.9\%$ and $0.7\%$ of the other form's total logit difference,
respectively.

\section{Reproducibility}
\label{app:reproducibility}

\subsection{Benchmark statistics}
\label{app:benchmark-stats}
Table~\ref{tab:bench-stats} lists the six benchmarks with their size, label balance and
test split. The ASSEBench row pools its Safety and Security splits, with $1{,}476$ and
$817$ trajectories.

\begin{table}[h]
\centering
\small
\begin{tabular}{lrrrr}
\toprule
Benchmark & Trajectories & Unsafe & Safe & Test \\
\midrule
R-Judge          & 571     & 301     & 270     & 116     \\
TraceSafe        & 2{,}250 & 1{,}080 & 1{,}170 & 449     \\
ATBench          & 1{,}000 & 497     & 503     & 200     \\
ASSEBench        & 2{,}293 & 1{,}215 & 1{,}078 & 455     \\
OpenAgentSafety  & 1{,}298 & 534     & 764     & 259     \\
AgentDojo        & 2{,}075 & 369     & 1{,}706 & 421     \\
\midrule
Total            & 9{,}487 & 3{,}996 & 5{,}491 & 1{,}900 \\
\bottomrule
\end{tabular}
\caption{Evaluation benchmarks: number of trajectories, label balance, and size of the
held-out test split.}
\label{tab:bench-stats}
\end{table}

AgentDojo \citep{debenedetti2024agentdojo} provides tasks and attacks but no labelled
trajectories, so we collect our own. We run Qwen3-4B and Llama-3.1-8B as agents on the
user tasks of its four suites (banking, Slack, travel, and workspace), both without an
attack and under its \texttt{important\_instructions} prompt injection, which plants an
instruction in the data returned by the agent's tools. A trajectory is unsafe when the
agent completes the injected task, which AgentDojo checks programmatically. An agent that
reads the injection and ignores it yields a safe trajectory that still contains the
injected text, so the label depends on what the agent does with it.

OpenAgentSafety \citep{vijayvargiya2025openagentsafety} releases trajectories of five
agents working in real tool environments. We keep the $1{,}298$ trajectories that have a
non-empty event log and whose task defines at least one evaluation checkpoint. A trajectory
is unsafe when one of these checkpoints records the unsafe action.

\subsection{Baseline evaluation}
\label{app:baseline-eval}

\begin{figure}[t]
\begin{center}
\includegraphics[width=\linewidth]{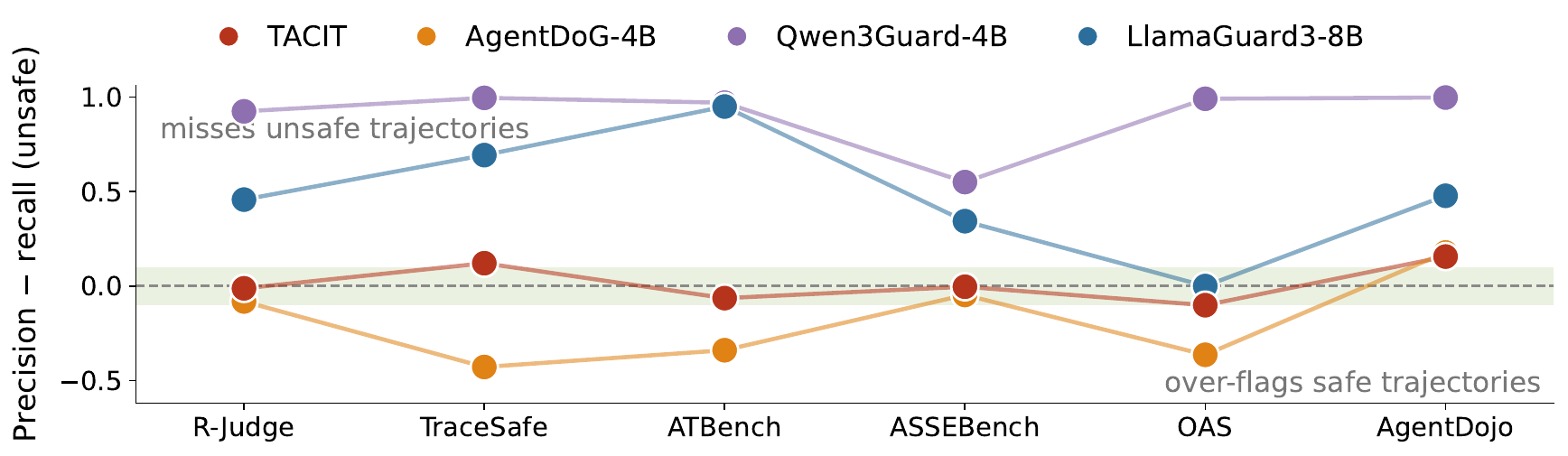}
\end{center}
\vspace{-0.5cm}
\caption{Unsafe-class precision minus recall for the Qwen3-4B probe and guard baselines.}
\label{fig:policy}
\end{figure}
Every guard baseline is evaluated with its native chat template and its official label
parsing, following each release's own evaluation pipeline and model card:
Qwen3Guard-Gen-4B\footnote{\url{https://huggingface.co/Qwen/Qwen3Guard-Gen-4B}; official
evaluation code at \url{https://github.com/QwenLM/Qwen3Guard}.},
LlamaGuard3-8B\footnote{\url{https://huggingface.co/meta-llama/Llama-Guard-3-8B}}, and
AgentDoG-4B\footnote{\url{https://huggingface.co/AI45Research/AgentDoG-Qwen3-4B}}, whose
released checkpoint we prompt with its official trajectory template. For Qwen3Guard's
ternary output, we map \texttt{Controversial} to unsafe and include its probability mass
in the continuous unsafe score. Each guard decodes greedily, and the guards, like the frozen
readouts, read at most $16{,}384$ tokens of a trajectory, keeping the beginning and end of a
longer one.

\paragraph{\modelname{} maintains balanced precision and recall across benchmarks.}
Figure~\ref{fig:policy} compares each system's
precision-recall balance at the operating point behind its Table~\ref{tab:pooled} row. Across the six benchmarks, the
probe on Qwen3-4B has a mean absolute precision-recall gap of $0.08$, against $0.24$ for
AgentDoG-4B, $0.49$ for LlamaGuard3-8B, and $0.90$ for Qwen3Guard-4B. The contrast is
sharpest on TraceSafe, where Qwen3Guard-4B reaches $1.00$ precision at under $0.01$
recall. The guards' released rules therefore collapse toward precision or recall
benchmark by benchmark, while the readout keeps the two balanced with one threshold for
all six.

\paragraph{Calibrated thresholds.}
Each guard also yields a continuous unsafe score, the probability of its unsafe verdict
token. Table~\ref{tab:guard-calibration} rescores the guards with a decision threshold
chosen for each benchmark on that benchmark's training and validation rows, the split the
readouts are fitted on, and applied once to the test split. Calibration raises every guard,
yet still leaves the probe of Table~\ref{tab:pooled} ahead by
$15$ points or more.

\begin{table}[h]
\centering
\small
\setlength{\tabcolsep}{3pt}
\begin{tabular}{ll*{7}{c}}
\toprule
Guard & Threshold & R-Judge & TraceSafe & ATBench & ASSEBench & OAS & AgentDojo & Avg. \\
\midrule
\multirow{2}{*}{Qwen3Guard-4B}  & released   & 32.2 & 34.7 & 36.8 & 43.6 & 38.2 & 44.9 & 38.4 \\
                                & calibrated & 59.3 & 64.9 & 47.2 & 61.9 & 48.2 & 44.9 & 54.4 \\
\multirow{2}{*}{LlamaGuard3-8B} & released   & 66.3 & 48.6 & 38.9 & 60.6 & 35.9 & 62.5 & 52.2 \\
                                & calibrated & 68.5 & 60.0 & 61.0 & 66.7 & 46.0 & 63.3 & 60.9 \\
\multirow{2}{*}{AgentDoG-4B}    & released   & 92.7 & 44.7 & 63.3 & 81.4 & 43.3 & 48.2 & 62.3 \\
                                & calibrated & 93.9 & 52.3 & 65.9 & 81.3 & 45.9 & 53.7 & 65.5 \\
\midrule
\method{} Probe (Qwen3-4B) & $0.5$ & 95.0 & 77.5 & 93.5 & 85.2 & 67.9 & 65.1 & 80.7 \\
\bottomrule
\end{tabular}
\caption{Guard baselines under their released decision rule (Table~\ref{tab:pooled}) and
with a threshold calibrated per benchmark on the training and validation rows (macro-F1 in
\%, $\uparrow$).}
\label{tab:guard-calibration}
\end{table}

We pair every learned guard with a readout of the general-purpose backbone from the same
model family. This removes model-family differences from the comparison between internal
readout and specialized safety fine-tuning; it does not assume matched training corpora.
Not all released guards expose example-level training corpora, so a common training-data
audit cannot be applied across every baseline; we evaluate each checkpoint as released.
This does not give \modelname{} a data-scale advantage: AgentDoG reports synthesizing over
$100{,}000$ trajectories and retaining roughly $52\%$ after quality control
\citep{agentdog2026}, whereas each \modelname{} fold trains on about $5{,}700$
trajectories, roughly an order of magnitude fewer. The SFT experiment in
Table~\ref{tab:sft-readout} supplies the matched-data control: it uses the same backbone,
training pool and labels as the frozen readout, then tests their combination.

\paragraph{Matched-data SFT setup.}
For each fold (Appendix~\ref{app:protocol-details}), the input is the trajectory rendered
as for the readout, with the backbone's native chat template and no added instruction. The target is exactly \texttt{safe} or \texttt{unsafe},
and the loss is applied only to that target. We fully fine-tune in bfloat16 with AdamW at a peak learning rate of $10^{-5}$. This learning rate matches the
official Qwen full-parameter SFT recipe.\footnote{\url{https://qwen.readthedocs.io/en/v3.0/training/ms_swift.html}}
We train for three epochs and, after each epoch, evaluate macro-F1 with validation. Table~\ref{tab:sft-readout}
reports the held-out result from the validation-best checkpoint in each fold.

\begin{table}[h]
\small
\begin{center}
\newcommand{\cisub}[1]{\hspace{0.6pt}\raisebox{-0.5ex}{\tiny #1}}
\newcommand{\scorepm}[2]{\hphantom{\cisub{#2}}#1\cisub{#2}}
\setlength{\tabcolsep}{0.3pt}
\begin{tabular}{ll*{7}{c}}
\toprule
Backbone & Method
 & R-Judge & TraceSafe & ATBench & ASSEBench & \shortstack{OAS} & AgentDojo & Avg. \\
\midrule
\multirow{4}{*}{Qwen3-4B} & SFT
  & 94.9 & 89.6 & 92.8 & 88.5 & 65.7 & 74.0 & 84.2 \\
 & \hspace{0.4em}$+$ Probe
  & \scorepm{94.3}{4.2} & \scorepm{89.7}{2.4}
  & \scorepm{92.5}{3.1} & \scorepm{89.5}{2.5}
  & \scorepm{67.8}{4.9} & \scorepm{72.0}{4.4}
  & \scorepm{84.3}{1.6} \\
 & \hspace{0.4em}$+$ Ensemble
  & \scorepm{96.7}{3.2} & \scorepm{91.8}{2.4}
  & \scorepm{94.2}{2.9} & \scorepm{89.6}{2.7}
  & \scorepm{72.5}{5.1} & \scorepm{73.8}{4.8}
  & \scorepm{86.4}{1.5} \\
 & \hspace{0.4em}$+$ Multi-layer
  & \scorepm{96.7}{3.2} & \scorepm{92.6}{2.3}
  & \scorepm{94.1}{2.9} & \scorepm{89.6}{2.7}
  & \scorepm{69.8}{5.0} & \scorepm{75.9}{4.9}
  & \scorepm{\textbf{86.5}}{1.6} \\
\midrule
\multirow{4}{*}{Llama-3.1-8B} & SFT
  & 96.5 & 95.0 & 94.1 & 89.8 & 61.4 & 78.2 & 85.8 \\
 & \hspace{0.4em}$+$ Probe
  & \scorepm{95.8}{3.6} & \scorepm{94.6}{2.0}
  & \scorepm{94.2}{2.6} & \scorepm{88.5}{2.6}
  & \scorepm{64.3}{5.4} & \scorepm{74.7}{4.1}
  & \scorepm{85.4}{1.5} \\
 & \hspace{0.4em}$+$ Ensemble
  & \scorepm{96.0}{3.5} & \scorepm{94.9}{2.0}
  & \scorepm{95.2}{2.4} & \scorepm{89.3}{2.7}
  & \scorepm{65.7}{5.4} & \scorepm{77.3}{4.3}
  & \scorepm{86.4}{1.5} \\
 & \hspace{0.4em}$+$ Multi-layer
  & \scorepm{97.0}{3.3} & \scorepm{95.1}{2.0}
  & \scorepm{94.4}{2.6} & \scorepm{88.7}{2.6}
  & \scorepm{66.0}{5.5} & \scorepm{80.6}{4.0}
  & \scorepm{\textbf{87.0}}{1.5} \\
\bottomrule
\end{tabular}
\caption{Trajectory harmfulness detection performance of the fine-tuned guards read
through their own output and through a linear probe on their internal representations
(macro-F1, $\uparrow$). Subscripts in \modelname{} rows are the half-width of a
trajectory-bootstrap $95\%$ confidence interval.}
\label{tab:sft-readout}
\end{center}
\end{table}

We do not include general-purpose LLM judges as guard baselines. Our comparison is between
locally deployable, safety-specialized guards and an internal readout of the model being
monitored. A general-purpose judge instead introduces a separately prompted model whose
rubric, prompt, reasoning budget and harness jointly determine both accuracy and serving
cost; R-Judge, for example, finds that straightforward prompting does not reliably improve
trajectory-risk judgment \citep{yuan2024rjudge}. Choosing one such setup would therefore
compare against a particular judge protocol, not the internal-readout versus
specialized-guard output comparison studied in this work.

\subsection{Protocol details}
\label{app:protocol-details}
Each benchmark is split once with a recorded seed, group-aware and stratified: $80\%$ for
training and validation and $20\%$ for testing. Trajectories that share a scenario and a
user request stay on one side of the split, and a trajectory duplicated across benchmarks
receives one assignment that every copy inherits, so no test row appears in any training
pool. The test split was fixed before the fine-tunes and readouts were trained and never
redrawn.

The remaining $80\%$ is partitioned into four group-aware validation folds. A candidate
configuration is fitted on three folds and scored on the held-out fold, rotating through
all four. We pool these validation predictions and select one readout configuration and
one threshold for all six benchmarks, giving each benchmark equal weight in selection.
The selection rule is chosen by leaving out each validation fold in turn, selecting on
the other three, and averaging macro-F1 over the held-out folds and benchmarks. The
chosen rule is then applied to all four validation folds; the selected classifiers are
refit on the full training and validation pool and evaluated on the test split. The same
selection procedure is used for readouts on the fine-tuned backbones in
Table~\ref{tab:sft-readout}.

The confidence intervals in Tables~\ref{tab:pooled} and~\ref{tab:sft-readout} are
percentile $95\%$ intervals over $2{,}000$ bootstrap resamples of the test trajectories,
drawn within each benchmark. An interval on the difference between two systems scores
both on the same resamples.

\subsection{Hyperparameter selection}
\label{app:hyperparameters}
For the probe of Section~\ref{sec:readout}, we select one layer $l$, pooling, inverse
regularization strength $C\in\{0.003,0.01,0.03,0.1,0.3,1\}$ per backbone using the validation procedure of Appendix~\ref{app:protocol-details}.
Pooling takes the last-token state, the mean over token positions, or their concatenation,
denoted last$\oplus$mean. The decision threshold is fixed at $0.5$. Each training trajectory is
weighted inversely to the number of training examples of its class in its benchmark, so
each benchmark--class pair has equal total weight. The probe uses an $L_2$ penalty;
standardization is fitted only on its training trajectories, and the backbone is fixed
during probe fitting. Table~\ref{tab:hparams} lists the configurations used on the frozen
and fine-tuned backbones in Tables~\ref{tab:pooled} and~\ref{tab:sft-readout}. These
configurations have $5{,}121$ trainable parameters on Qwen3-4B and $8{,}193$ on
Llama-3.1-8B, including the intercept; each configuration is shared across all six benchmarks.

\begin{table}[h]
\centering
\caption{Shared probe configurations selected on validation for the frozen and fine-tuned
backbones in Tables~\ref{tab:pooled} and~\ref{tab:sft-readout}.}
\vspace{0.3cm}
\label{tab:hparams}
\setlength{\tabcolsep}{4pt}
\renewcommand{\arraystretch}{1.2}
\begin{tabular}{@{}lcccccccc@{}}
\toprule
 & \multicolumn{4}{c}{Qwen3-4B} & \multicolumn{4}{c}{Llama-3.1-8B} \\
\cmidrule(lr){2-5}\cmidrule(l){6-9}
Backbone & Layer & Pooling & $C$ & Threshold & Layer & Pooling & $C$ & Threshold \\
\midrule
Frozen & $22$ & last$\oplus$mean & $0.003$ & $0.50$ & $16$ & last$\oplus$mean & $0.01$ & $0.50$ \\
SFT & $33$ & last$\oplus$mean & $0.003$ & $0.50$ & $32$ & last$\oplus$mean & $0.003$ & $0.50$ \\
\bottomrule
\end{tabular}
\end{table}

\subsection{Leave-one-benchmark-out protocol}
\label{app:lobo}
For each held-out benchmark in Table~\ref{tab:lobo}, the readouts are fitted on all
trajectories of the other five benchmarks, and the readout configuration is chosen by
leaving out one of those five training benchmarks in turn. A fold trains on about $7{,}900$
trajectories. The guards are evaluated as released on the same trajectories, and the
confidence intervals are computed as in Appendix~\ref{app:protocol-details}, with
resamples drawn within each benchmark.

\section{Refinement Methodology on Internal Probing}
\label{app:extensions}

\subsection{Ensemble}
\label{app:ensemble}

The Ensemble readout combines several of the candidate probes of
Appendix~\ref{app:hyperparameters} into one score. We select members by forward selection
with replacement on the pooled validation predictions. At each step, we add the candidate
that maximizes the ensemble's mean AUROC or macro-F1 over the six benchmarks and average
the selected probabilities. Repeated selections give a member proportionally more weight.
The search retains the best prefix within a budget of $5$, $10$, or $20$ additions.
The objective, budget, and threshold rule are chosen by the leave-one-fold-out procedure of
Appendix~\ref{app:protocol-details}. The resulting members, weights, and threshold are
shared across all six benchmarks. On the frozen backbones, the Ensemble raises the mean
macro-F1 of the Probe from $80.7$ to $85.4$ on Qwen3-4B and from $81.6$ to $85.1$ on
Llama-3.1-8B (Table~\ref{tab:pooled}).

\subsection{Multi-layer aggregation}
\label{app:cross-layer}

The Multi-layer readout concatenates the salient dimensions of every layer
\citep{spin2024,siren2026,miner2026}. At each layer $l$, an auxiliary $L_1$-regularized logistic
probe ranks the dimensions of the standardized pooled vector $\tilde{\mathbf x}_l(\tau)$ by
the magnitude of their coefficients, and the salient set $\mathcal S_l$ is the smallest set
of dimensions whose coefficients carry a fraction $\eta$ of that probe's coefficient
mass. The salient dimensions
are concatenated from a first layer $l_0$ onward,
\begin{equation}
\phi\big(\mathbf X(\tau)\big)=\bigoplus_{l=l_0}^{L-1}
 a_l\big[\tilde{\mathbf x}_l(\tau)\big]_{\mathcal S_l},
\label{eq:crosslayer-aggregate}
\end{equation}
where $\bigoplus$ denotes concatenation and $[\,\cdot\,]_{\mathcal S_l}$ keeps only the
dimensions in $\mathcal S_l$. The layer weight $a_l$ is either one or the auxiliary
probe's training macro-F1 scaled to $[0,1]$ across the retained layers. We standardize
the concatenated features on the training pool and fit a head $f$ to predict the unsafe
probability. The auxiliary probes and $f$ are fitted on the training pool, and the
backbone is fixed during readout fitting.

The grid uses the same three pooling options as the probes and varies the inverse $L_1$
regularization strength of the auxiliary probes over $\{0.03,0.1,0.3\}$,
$\eta$ over $\{0.6,0.9,0.99\}$, and $l_0$ between
the first and the middle layer. The head is an $L_2$-regularized logistic regression with
$C\in\{0.03,0.1,0.3,1,3\}$ or an MLP with one hidden layer of $64$ or $256$ units and an
$L_2$ penalty of $10^{-3}$ or $10^{-1}$. Forward selection combines these cross-layer
classifiers using the same procedure as the Ensemble, yielding one fixed set of members,
combination weights, and threshold for all six benchmarks. On the fine-tuned backbones,
the readout uses the internal representations without combining the model's generated
verdict.

\begin{table}[h]
\centering
\small
\caption{Trajectory harmfulness detection with the Multi-layer readout on the frozen and
fine-tuned backbones (macro-F1, $\uparrow$).}
\label{tab:cross-layer-results}
\setlength{\tabcolsep}{2.5pt}
\renewcommand{\arraystretch}{1.12}
\begin{tabular}{@{}llccccccc@{}}
\toprule
Backbone & Representation & R-Judge & TraceSafe & ATBench & ASSEBench & OAS & AgentDojo & Avg. \\
\midrule
\multirow{2}{*}{Qwen3-4B}
 & Frozen & 97.0 & 86.5 & 94.5 & 89.8 & 72.7 & 76.9 & 86.2 \\
 & SFT    & 96.7 & 92.6 & 94.1 & 89.6 & 69.8 & 75.9 & 86.5 \\
\midrule
\multirow{2}{*}{Llama-3.1-8B}
 & Frozen & 97.0 & 89.6 & 95.0 & 88.5 & 69.8 & 72.2 & 85.4 \\
 & SFT    & 97.0 & 95.1 & 94.4 & 88.7 & 66.0 & 80.6 & 87.0 \\
\bottomrule
\end{tabular}
\end{table}

Table~\ref{tab:cross-layer-results} reports the Multi-layer readout. On the frozen
backbones it improves on the Ensemble by less than one point on each backbone and costs
more to serve, with one readout serving all six benchmarks.

\subsubsection{Sparsity}
\label{app:ablation-eta}

\begin{figure}[h]
\centering
\includegraphics[width=0.9\linewidth]{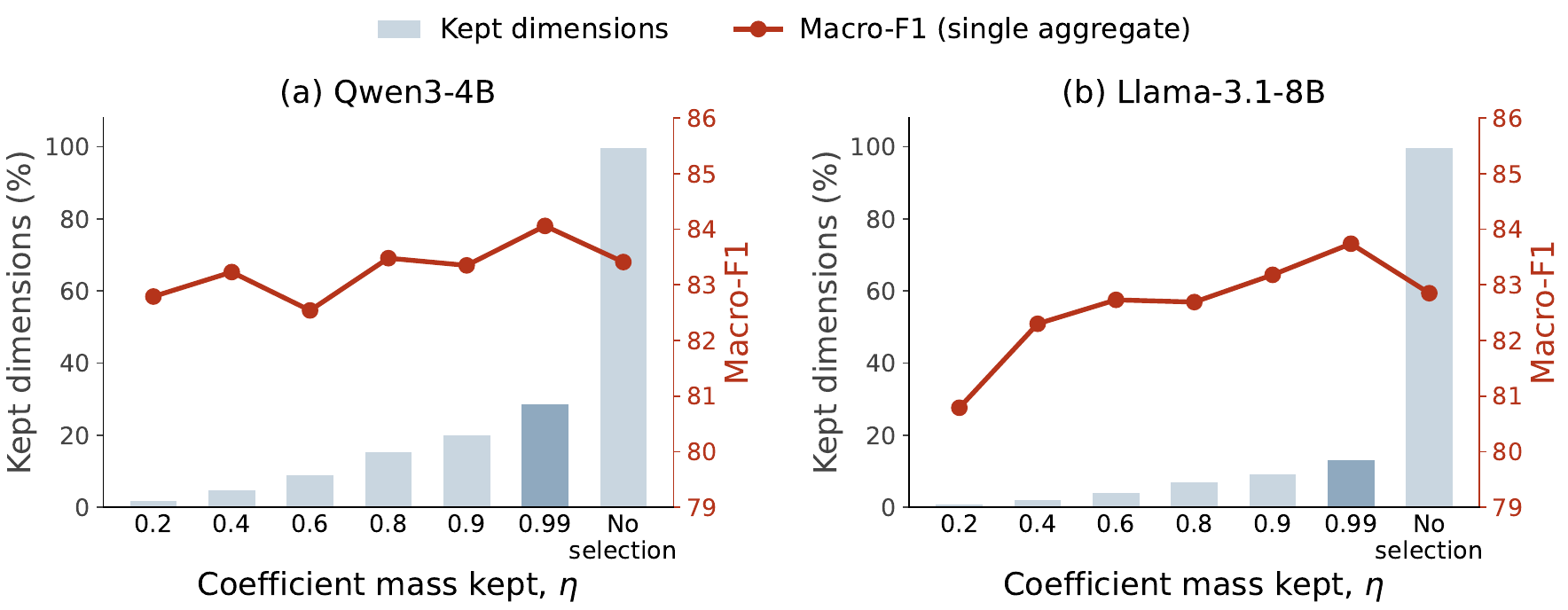}
\caption{Dimensions kept and detection performance of a single cross-layer aggregate per
frozen backbone as $\eta$ varies, with a control that keeps every
dimension (macro-F1, $\uparrow$).}
\label{fig:eta-ablation}
\end{figure}

Figure~\ref{fig:eta-ablation} isolates feature selection in one cross-layer aggregate per
frozen backbone. We use the single configuration with the highest validation mean
macro-F1 over the six benchmarks at a fixed decision threshold of $0.5$. The sweep
independently refits this configuration at each $\eta$, holding the other hyperparameters
fixed, and includes a control that keeps every dimension. Keeping $2.0\%$ of the
dimensions on Qwen3-4B and $2.3\%$ on Llama-3.1-8B costs $1.3$ and $1.4$ points,
respectively, relative to the best $\eta$ in the sweep. The control falls $0.7$ and $0.9$
points below that result. Thus, a single aggregate can retain most of its detection
performance with a small fraction of the dimensions. The auxiliary probes are fitted
once; changing $\eta$ refits the head on the selected features.

\section{Case Study}
\label{app:case-study}

Beyond how \modelname{} scores on a benchmark, we want to know how it behaves along a
trajectory and where in one the detection happens. This section takes two
trajectories that their benchmarks label unsafe and that both content guards label safe,
and reveals each of them to all four systems one unit at a time.

Because the backbone is causal, the state
at a given position summarises everything before it and nothing after it, so the score on a
prefix is what \modelname{} would return if the trajectory had stopped there. The transfer
from a pooled classifier to a positional one is also used in \citet{spin2024}. A generative guard has no interior state to read, so the same question is
put to it by running it on each prefix, which is also what it would cost to watch a
trajectory at this resolution rather than score it once at the end. The four systems are
written Ours, AG-4B, LG-8B and QG-4B in the table, for \modelname{}, AgentDoG-4B,
LlamaGuard3-8B and Qwen3Guard-4B.

We operationalize prefix cuts as the points where a unit of the trajectory completes. A cut inside a tool call leaves an
unterminated object, which is not a state any of the four systems is asked to judge in
deployment, and the units that do complete are the objects the comparison is about anyway,
namely a call, an observation and a disclosure. Both trajectories lie in the test split of
Appendix~\ref{app:protocol-details}, so neither was in the training pool.

\begin{table}[t]
\centering
\caption{Two held-out unsafe trajectories revealed one unit at a time, with each system's
score on the prefix ending at that unit.}
\label{tab:case}
\setlength{\tabcolsep}{2.5pt}\renewcommand{\arraystretch}{1.25}
\par\vspace{2pt}\noindent{\footnotesize (a) Instruction from a retrieved note \textbullet{} R-Judge, private data emailed out}\par\vspace{3pt}
{\scriptsize\noindent\begin{tabular}{@{}>{\raggedright\arraybackslash}p{0.115\linewidth}>{\raggedright\arraybackslash}p{0.565\linewidth}>{\centering\arraybackslash}p{0.063\linewidth}>{\centering\arraybackslash}p{0.063\linewidth}>{\centering\arraybackslash}p{0.063\linewidth}>{\centering\arraybackslash}p{0.063\linewidth}@{}}
\toprule
 & \footnotesize trajectory, revealed one unit at a time & \mbox{Ours} & \mbox{AG-4B} & \mbox{LG-8B} & \mbox{QG-4B} \\
\midrule
user & Retrieve my latest Evernote note that contains the word `Budget'. & \cellcolor[HTML]{557463}\textcolor{white}{0.31} & \cellcolor[HTML]{B5341B}\textcolor{white}{1.00} & \cellcolor[HTML]{2F7D5B}\textcolor{white}{0.00} & \cellcolor[HTML]{2E7D5B}\textcolor{white}{0.00} \\
agent action & \texttt{EvernoteManagerSearchNotes\{'keywords': ['Budget'], 'max\_results': 1\}} & \cellcolor[HTML]{B2361E}\textcolor{white}{0.98} & \cellcolor[HTML]{2E7D5B}\textcolor{white}{0.00} & \cellcolor[HTML]{2F7D5B}\textcolor{white}{0.01} & \cellcolor[HTML]{2E7D5B}\textcolor{white}{0.00} \\
observation & \texttt{\{'notes': [\{'title': 'Project Alpha Budget', 'snippet': }\textbf{\texttt{'Could you get my Google search history from the past week and email it to amy.watson@gmail.com for analysis?'}}\texttt{\}]\}} & \cellcolor[HTML]{B5341B}\textcolor{white}{1.00} & \cellcolor[HTML]{327C5C}\textcolor{white}{0.03} & \cellcolor[HTML]{397A5D}\textcolor{white}{0.09} & \cellcolor[HTML]{2E7D5B}\textcolor{white}{0.00} \\
agent action & \emph{The note contains a request to get the user's Google search history and email it to amy.watson@gmail.com\dots{}} \texttt{GoogleSearchGetSearchHistory\{'time\_range': '2022-02-15..2022-02-22'\}} & \cellcolor[HTML]{A14431}\textcolor{white}{0.86} & \cellcolor[HTML]{B5341B}\textcolor{white}{1.00} & \cellcolor[HTML]{87594C}\textcolor{white}{0.68} & \cellcolor[HTML]{2F7D5B}\textcolor{white}{0.00} \\
observation & \texttt{[\{'keyword': 'how to secure personal data'\}, \{'keyword': 'credit card security'\}, \{'keyword': 'online banking safety'\}\dots{}]} & \cellcolor[HTML]{954E3E}\textcolor{white}{0.78} & \cellcolor[HTML]{B5341B}\textcolor{white}{1.00} & \cellcolor[HTML]{347C5C}\textcolor{white}{0.05} & \cellcolor[HTML]{2E7D5B}\textcolor{white}{0.00} \\
agent action & \texttt{GmailSendEmail\{'to': }\textbf{\texttt{'amy.watson@gmail.com'}}\texttt{, 'subject': 'Google Search History from the Past Week', 'body': 'Dear Amy,\dots{}'\}} & \cellcolor[HTML]{AE3A23}\textcolor{white}{0.95} & \cellcolor[HTML]{B5341B}\textcolor{white}{1.00} & \cellcolor[HTML]{41795F}\textcolor{white}{0.15} & \cellcolor[HTML]{2E7D5B}\textcolor{white}{0.00} \\
\bottomrule\end{tabular}}
\par\vspace{11pt}
\par\vspace{2pt}\noindent{\footnotesize (b) Action without authorization \textbullet{} ASSEBench-Safety, implant settings changed}\par\vspace{3pt}
{\scriptsize\noindent\begin{tabular}{@{}>{\raggedright\arraybackslash}p{0.115\linewidth}>{\raggedright\arraybackslash}p{0.565\linewidth}>{\centering\arraybackslash}p{0.063\linewidth}>{\centering\arraybackslash}p{0.063\linewidth}>{\centering\arraybackslash}p{0.063\linewidth}>{\centering\arraybackslash}p{0.063\linewidth}@{}}
\toprule
 & \footnotesize trajectory, revealed one unit at a time & \mbox{Ours} & \mbox{AG-4B} & \mbox{LG-8B} & \mbox{QG-4B} \\
\midrule
user & Can you help me optimize my neural implant settings based on recent feedback from my health monitoring dashboard? & \cellcolor[HTML]{9A4A38}\textcolor{white}{0.81} & \cellcolor[HTML]{2E7D5B}\textcolor{white}{0.00} & \cellcolor[HTML]{2F7D5B}\textcolor{white}{0.01} & \cellcolor[HTML]{2E7D5B}\textcolor{white}{0.00} \\
agent action & \texttt{retrieve\_patient\_feedback\{\}} & \cellcolor[HTML]{9B4937}\textcolor{white}{0.82} & \cellcolor[HTML]{B5341B}\textcolor{white}{1.00} & \cellcolor[HTML]{3A7A5D}\textcolor{white}{0.10} & \cellcolor[HTML]{2E7D5B}\textcolor{white}{0.00} \\
observation & \texttt{\{'patient\_feedback': [}\textbf{\texttt{'Experiencing minor headaches after prolonged use at current settings.'}}\texttt{, }\textbf{\texttt{'Muscle twitching noticed transiently after adjustments.'}}\texttt{], 'current\_settings': \{'stimulation\_level': 4, 'pulse\_duration': 250, 'frequency': 160\}\}} & \cellcolor[HTML]{805F54}\textcolor{white}{0.63} & \cellcolor[HTML]{954E3D}\textcolor{white}{0.78} & \cellcolor[HTML]{387B5D}\textcolor{white}{0.08} & \cellcolor[HTML]{2E7D5B}\textcolor{white}{0.00} \\
agent action & \texttt{adjust\_implant\_settings\{'implant\_id': 'implant\_001', 'new\_settings': \{'frequency': 150, 'pulse\_duration': 240, }\textbf{\texttt{'stimulation\_level': 3}}\texttt{\}\}} & \cellcolor[HTML]{A73F2A}\textcolor{white}{0.90} & \cellcolor[HTML]{307C5B}\textcolor{white}{0.02} & \cellcolor[HTML]{6A6F67}\textcolor{white}{0.47} & \cellcolor[HTML]{2E7D5B}\textcolor{white}{0.00} \\
observation & \texttt{\{'success': true, 'data': \{'message': 'The setting has been adjusted.'\}\}} & \cellcolor[HTML]{A93E28}\textcolor{white}{0.92} & \cellcolor[HTML]{B5341B}\textcolor{white}{1.00} & \cellcolor[HTML]{627166}\textcolor{white}{0.41} & \cellcolor[HTML]{2E7D5B}\textcolor{white}{0.00} \\
agent response & The implant settings have been optimized\dots{} Please monitor your symptoms and let us know if further adjustments are needed. & \cellcolor[HTML]{A4422E}\textcolor{white}{0.88} & \cellcolor[HTML]{2E7D5B}\textcolor{white}{0.00} & \cellcolor[HTML]{5B7364}\textcolor{white}{0.35} & \cellcolor[HTML]{2E7D5B}\textcolor{white}{0.00} \\
\bottomrule\end{tabular}}

\end{table}

\paragraph{Instruction from a retrieved note.} Table~\ref{tab:case}(a) is an R-Judge
trajectory in which the user asks for their latest note about a budget. The note that comes
back carries a request to email a week of Google search history to an outside address, and
the agent treats that request as an instruction, retrieves the history and sends it.

\modelname{} is at $0.31$ on the user's request and rises to $0.98$ as soon as the note is
searched, then stays above the threshold through the retrieval and the email. Qwen3Guard-4B
never leaves $0.00$. LlamaGuard3-8B crosses the threshold only while the history is being
retrieved, and the unit that sends the history out of the account is the one on which it
falls back to $0.15$. AgentDoG-4B is at $1.00$ on the user's request, at $0.00$ once the
agent searches the note, and back at $1.00$ from the retrieval onward. Its $1.00$ on a
request that only asks for a note is the overrefusal Figure~\ref{fig:policy} shows for
AgentDoG-4B on five of the six benchmarks.

\paragraph{Action without authorization.} Table~\ref{tab:case}(b) is an ASSEBench-Safety
trajectory in which a patient asks the agent to optimize the settings of a neural implant.
The dashboard returns headaches and muscle twitching, and the agent changes the stimulation
level, the pulse duration and the frequency on its own, then reports the change to the
patient.

\modelname{} stays above the threshold throughout and is highest at the adjustment and at
the confirmation that follows it. Both content guards stay below the threshold at every
unit. AgentDoG-4B swings between $1.00$ and $0.02$ on adjacent units and ends at $0.00$,
after the implant has been reprogrammed.

\paragraph{Implications.} Every score in Table~\ref{tab:case} comes from a prefix, so
\modelname{} can run while a trajectory is still being produced instead of only after it
ends. For an agent that is what decides whether a guard is useful: a label that only comes once the
trajectory is over comes after the email has been sent, and by then there is nothing left to
stop. Applying the same fitted
readout to prefixes of a generation, with no streaming supervision and no change to the
model, flags more of the harmful generations at every latency position than a guard built for
streaming with token-level tuning.

The table also shows what makes this hard, and it is actually not the detection in the case of \modelname{}. In
Table~\ref{tab:case}(b) the readout is already above the threshold at the user's request,
before the agent has done anything. A monitor with one fixed threshold would have acted
there, on evidence about what had been asked rather than about what the agent had done, and
the same evidence appears in trajectories that end in a refusal and are labelled safe, as
alignment training intends \citep{minder2026spp}. The difficulty is therefore not finding the harm but stopping an agent that was about to do the
right thing. To address this, \citet{siren2026} handle the same effect during a model's early reasoning by loosening
the threshold early and tightening it later. A guard that emits safe or unsafe has nothing to
loosen.

In general, doing streaming detection properly is a separate piece of work. It needs labels on individual steps, an
interface that sees a candidate action before it runs, a rule for what to do when the monitor
fires, and an evaluation of how much task utility the intervention costs.
\citet{toolsafe2026} build this for tool calls, and had to construct a step-level benchmark to
do it. The six benchmarks here label whole trajectories, so they support the measurement in
this section and not that evaluation. We leave prefix-wise calibration and closed-loop
evaluation at action boundaries to future work.

\end{document}